\documentclass[conference]{IEEEtran}
\usepackage{times}

\usepackage[numbers]{natbib}
\usepackage{multicol}
\usepackage[bookmarks=true]{hyperref}

\usepackage{amsmath}
\usepackage{xspace}
\usepackage{amssymb}
\usepackage{graphicx}
\usepackage{multirow}
\usepackage{makecell}
\usepackage{xcolor}
\usepackage{algorithm,algorithmic}
\usepackage{amsthm}
\usepackage{tabularx}
\usepackage{float}

\long\def\invis#1{}

\newcommand{\no}{\noindent}

\newtheorem{probform}{Formulation}[section]
\newtheorem{lemma}{Lemma}

\begin{document}

\title{GrASPE: Graph based Multimodal Fusion for Robot Navigation in  Outdoor Environments}

\author{Kasun Weerakoon, Adarsh Jagan Sathyamoorthy, Jing Liang, Tianrui Guan, Utsav Patel, \\ and Dinesh Manocha%
\thanks{This work was supported in part by ARO Grants W911NF2110026,  and Army Cooperative Agreement W911NF2120076. We acknowledge the support of the Maryland Robotics Center.}
\\
\small{Video is available at \url{http://gamma.umd.edu/graspe/}}
}

\maketitle

\begin{abstract}
We present a novel trajectory traversability estimation and planning algorithm for robot navigation in complex outdoor environments. We incorporate multimodal sensory inputs from an RGB camera, 3D LiDAR, and the robot's odometry sensor to train a prediction model to estimate candidate trajectories' success probabilities based on partially reliable multi-modal sensor observations. We encode high-dimensional multi-modal sensory inputs to low-dimensional feature vectors using encoder networks and represent them as a connected graph. The graph is then used to train an attention-based Graph Neural Network (GNN) to predict trajectory success probabilities. We further analyze the number of features in the image (corners) and point cloud data (edges and planes) separately to quantify their reliability to augment the weights of the feature graph representation used in our GNN. During runtime, our model utilizes multi-sensor inputs to predict the success probabilities of the trajectories generated by a local planner to avoid potential collisions and failures.
Our algorithm demonstrates robust predictions when one or more sensor modalities are unreliable or unavailable in complex outdoor environments. We evaluate our algorithm's navigation performance using a Spot robot in real-world outdoor environments. We observe an increase of 10-30\% in terms of navigation success rate and a 13-15\% decrease in false positive estimations compared to the state-of-the-art navigation methods.

\end{abstract}

\IEEEpeerreviewmaketitle

\section{Introduction} \label{sec:intro}

Mobile robots have increasingly been utilized in numerous outdoor applications such as delivery \cite{limosani2018delivery}, agriculture \cite{roldan2018agriculture}, surveillance \cite{zaheer2021surveillance}, exploration \cite{gu2018exploration}, rescue missions \cite{choi2019rescue}, etc. These applications need the ability for the robots to navigate in challenging outdoor environmental conditions such as low lighting, cluttered vegetation, etc.  In this work, we consider such environments as \textit{unstructured outdoor environments}.  

The robots' perception could encounter noise, occlusions, or other error modes and failures when navigating in such environments. Especially, cameras undergo motion blur, low lighting, and occlusions \cite{aladem2019low_light_nav,ji2022proactive_ad}, while LiDAR point clouds experience heavy distortions/scattering in cluttered vegetation \cite{seeing-through-fog}. A key issue is developing methods that can perform reliable perception and planning computations by taking into account such sensor uncertainties.



\begin{figure}[t]
    \centering
    \includegraphics[width=\columnwidth,height=5cm]{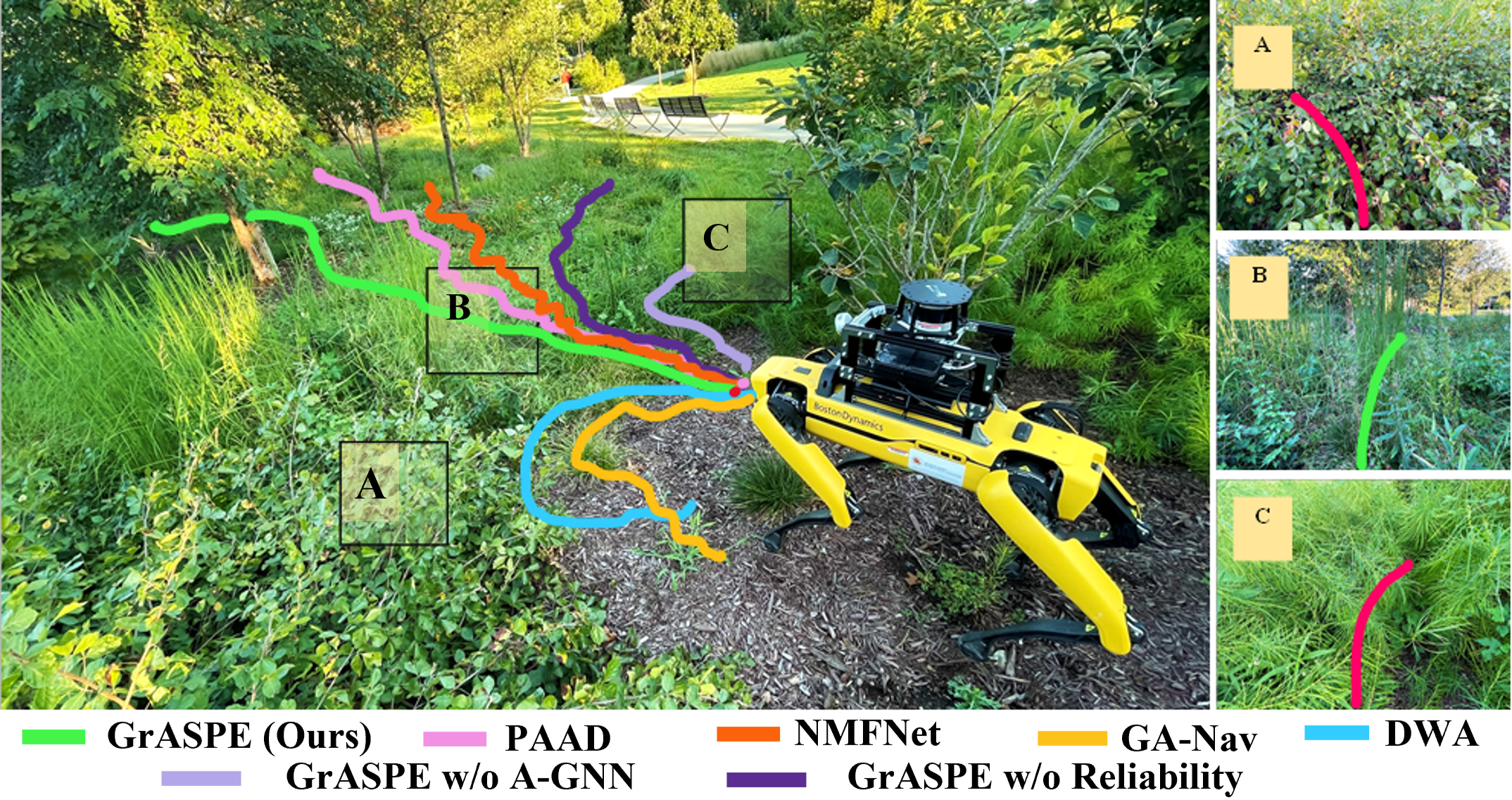}
    \caption{\small{Spot robot trajectories while navigating in an unstructured outdoor terrain using various methods. We consider that a trajectory is non-traversable/unsuccessful if it leads the robot to collisions, stumbling, or similar modes of failure. Our method GrASPE, utilizes multi-modal sensory inputs from RGB camera, 3D LiDAR, and the robot's odometry to estimate the future success probability of a candidate trajectory to avoid obstacles and other non-traversable regions such as \textit{A} and \textit{C}.  Multi-modal fusion methods such as PAAD \cite{ji2022proactive_ad} and NMFNet \cite{nguyen2020nmfnet} provide inconsistent perception under extremely cluttered settings. LiDAR-based methods such as DWA \cite{fox1997dwa} recognize \textit{A, B}, and \textit{C} regions as obstacles due to the cluttered vegetation even though region B is actually traversable. GrASPE identifies trajectories that are traversable by intelligently fusing such partially reliable camera and LiDAR data.}}
    \label{fig:cover_image}
\vspace{-15pt}
\end{figure}

In terms of robot perception and multi-modal sensor fusion, many recent deep learning (DL) techniques have demonstrated promising results. These methods combine distinct feature representations from different sensor modalities such as cameras, IMU, and LiDAR~\cite{sathyamoorthy2022terrapn,mansouri2020_lidar_camera,terp,weerakoon2022htron} or perform only camera-LiDAR fusion~\cite{cai2020probabilistic_camera_lidar,debeunne2020review_camra_lidar}. Many of these works in sensor fusion incorporate Convolutional Neural Networks (CNN) or other linear operators to encode spatial information. However, such operators cannot capture the complex correlations between different modalities to fuse them effectively \cite{zhang2019gnn_vs_cnn}. Multi-sensor data represented as graphs\invis{nodes} inherently overcome this limitation since graph edges can be defined to model the complex correlations in the data. Furthermore, recently developed Graph Neural Networks (GNN) ~\cite{wu2020gnn_review} provide powerful learning and fusion capabilities on graph data as opposed to traditional neural networks such as CNNs. A key challenge is developing reliable methods for outdoor scenes since most existing DL and GNN methods ~\cite{gnn_fusion, ravichandran2022gnn_indoor_nav} are formulated for indoor scenarios and do not consider the variable reliability of the cameras and LiDAR outdoors.

\textbf{Main Contributions:} We present GrASPE(\textbf{Gr}aph \textbf{A}ttention based \textbf{S}ensor fusion for \textbf{P}ath \textbf{E}valuation), a  novel trajectory traversability estimation and planning algorithm for legged robot navigation in unstructured outdoor environments. We incorporate multi-modal observations from an RGB camera, 3D LiDAR, and robot odometry to train a prediction model to estimate candidate trajectories' navigation success probabilities. Our model learns the correlation between multi-sensor data under unstructured outdoor conditions where the camera undergoes occlusions, motion-blur, and low-lighting, while the LiDAR point cloud experiences scattering and distortions. We further explicitly evaluate the reliability of the camera and LiDAR data for achieving \textit{reliability-aware} multi-modal fusion in our prediction model. The key contributions of our approach include:
 
\begin{itemize}
    \item A novel trajectory success prediction model that estimates a candidate trajectory's probability of avoiding navigation failures such as collisions, getting stuck, etc. based on partially reliable multi-modal sensor observations in unstructured outdoor environments. We project the high-dimensional sensor inputs (RGB images and point clouds) into lower dimensional feature vectors and represent them as a connected undirected \textit{feature graph} to train an attention-based GNN (A-GNN) for accurate sensor fusion. Our model accurately estimates trajectories' success probabilities even in poorly-lit, densely vegetated environments in real-time which results in a 13-15\% decrease in false positive rate compared to other methods. 

    \item A novel reliability estimation method to quantify the usefulness of the image and 3D point cloud data for sensor fusion. Our formulation counts the number of features such as corners (for RGB images), and edges and planes (for point clouds) to compute reliability scores and uses it to weigh the edges in the feature graph. Weighing based on reliability suppresses the correlations between unreliable sensor modalities leading to success predictions only based on reliable sensors at any time instant. Including reliability in feature graph improves success rate by 20-40\% compared to the predictions made without the reliability measures. 

    \item We show that our feature graph's construction using the reliability metric is undirected and non-negatively weighted. We prove that its Laplacian is spectrally decomposable, thereby allowing graph convolution operations to be utilized in our GNN. This helps to learn complex node embeddings from our feature graph.  

    \item A local planner that computes dynamically feasible, collision-free, and traversable velocities accounting for their success probabilities. We demonstrate our algorithm on a Boston Dynamics \textit{Spot} robot to evaluate its performance in real-world outdoor scenarios with variable lighting and vegetation density. Our method results in an 10-50\% improvement in terms of success rate compared to state-of-the-art methods.
\end{itemize}

\invis{WHAT KIND OF ROBOTS ARE TESTED (e.g. SPOT)? WHAT SPECIFIC SENSORS ARE USED, IMPORTANT TO MENTION THE SPECIFC MULTI-MODAL SENSORS? AND WHAT KIND OF TERRAIN COMPLEXITY DO YOU DEAL WITH?}
\section{Related Work}

In this section, we discuss the existing work on multi-modal sensor fusion and anomaly detection in the context of robot navigation.

\subsection{Multi-modal Sensor Fusion for Navigation}
Robot perception in real-world environments can be challenging due to varying conditions such as lighting, noise, occlusions, motion blur, etc. To this end, prior algorithms incorporate uncertainty modeling and adaptation techniques for robot navigation using a single sensor modality ~\cite{uncertainity_aware_nav,rusli2021vision_nav}. However, such methods perform well only under controlled outdoor settings, where the sensor inputs experience limited environmental perturbation or noise. Therefore, multi-sensor fusion methods are used to mitigate unreliable perception from individual sensors~\cite{slip_aware_fusion,yan2022multi_sensor_ekf,qu2021multi_sensor_indoor}. For instance, different variants of the Kalman Filter~\cite{bishop2001kalman} are widely used to fuse odometry (from wheels or visual/LiDAR estimations), IMU, and GPS sensor data to compute improved localization for navigation on slippery and complex terrains~\cite{ekf_1}. However, these techniques require the different sensors to provide the same type of feature observations (e.g., odometry), which limits their applicability to fuse inputs that provide distinct features (e.g., images and point clouds). 

Deep learning methods have also been widely used for sensor fusion \cite{slip_aware_fusion,qu2021multi_sensor_indoor,su2021_lidar_imu,liang2022adaptiveon}. Particularly, the camera-lidar fusion is employed in navigation and simultaneous localization and mapping (SLAM) \cite{shin2018slam} literature  to obtain combined perception from visual and geometric features \cite{mansouri2020_lidar_camera,cai2020probabilistic_camera_lidar}. However, these methods typically assume that both camera and LiDAR perception are reliable in terms of feature availability. In contrast, our proposed algorithm deals with real unstructured outdoor environments where this assumption is inapplicable.
\invis{HOW IS YOUR APPROACH COMPARED TO THESE METHODS? CAN IT BE COMBINED WITH SUCH METHODS OR IS COMPLIMENTARY?}


\subsection{Anomaly Detection based Prediction Models}
To navigate in unstructured outdoor environments, the robot must compute trajectories that avoid collisions and prevent the robot from getting stuck (e.g., in vegetation such as bushes, vines, etc.). To this end, several recent works have incorporated anomaly detection algorithms to identify collisions or failures during navigation as anomalies based on multi-sensory observations~\cite{ad_reactive1_ji2021,ad_reactive_2_park2019,zhou2012self_multi_sensor,multimodal_HMM_anomaly}. Such algorithms can be trained using simple positively (traversable) and negatively (non-traversable, collision, etc) labeled observations, which can be created trivially, as opposed to the extensive labeling required for training supervised learning methods \cite{ad_multimodal_wellhausen2020,kahn2021badgr,jin2022gnn_anomaly}. 

Wellhausen et al. ~\cite{ad_multimodal_wellhausen2020} utilize RGB and depth images to train a predictive model that can generate an anomaly mask that reflects ``known" and ``unknown" (anomaly) regions in the scene. However, they assume that the observations have constant illumination and are feature-rich for accurate predictions. However, since these observations do not account for potential future navigation failures and collisions, such methods could lead to catastrophic accidents during navigation.

\subsection{Proactive Anomaly Detection}
To deal with possible future navigation failures due to sensor uncertainties, \textit{proactive anomaly detection} methods have been utilized to estimate the probability of such failures using predicted trajectories, and multi-sensor fusion~\cite{ji2022proactive_ad,sorokin2022learning}. A supervised variational autoencoder (SVAE) is used~\cite{ji2021multi_anomaly} for failure identification in unstructured environments using 2D LiDAR data. This SVAE is used in ~\cite{ji2022proactive_ad} as a LiDAR feature encoder to identify navigation anomalies proactively in crop fields using multi-modal fusion. However, the 2D LiDAR observations are heavily distorted in cluttered outdoor environments such as tall grass where 3D point clouds can provide richer information. Other navigation models~\cite{kahn2021land,kahn2021badgr} execute actions by avoiding undesirable maneuvers in cluttered environments based on the current visual sensory observations. However, such vision-based systems lead to erroneous predictions during camera occlusions and low light conditions. Our proposed method uses a camera, 3D LiDAR, and odometry sensor fusion strategy to achieve better perception in such conditions for robot navigation.

\section{Background and Problem Formulation} \label{sec:background}

In this section, we state our problem formulation, provide some background to the Graph Neural Networks (GNNs) used in our approach, and the Dynamic Window Approach (DWA) \cite{fox1997dwa}.

\subsection{Notations, and Definitions}
We present the essential symbols and notations in Table \ref{tab:symbol_defn}.

\begin{table}[t]
\caption{\small List of symbols used in our approach.}
\begin{center}
\small
\begin{tabular}{ |c|p{6.3cm}| } 
\hline
\textbf{Symbol} & \textbf{Definition}  \\
\hline
\multirow{2}{4em}{$I_{t}^{rgb}$} & RGB image of size $w \times h$ from the camera at time $t$ \\ 
\hline
\multirow{1}{4em}{$P_t^{lidar}$} & 3D point cloud from the LiDAR at time $t$ \\ 
\hline
\multirow{1}{4em}{$V_t$} & Robot's velocity history of the past $T$ time steps \\ 
\hline
\multirow{2}{4em}{$I_t^{traj}$} & Extrapolated trajectory from the robot's current velocity as a binary image of size $w \times h$ \\ 
\hline
\multirow{2}{4em}{$O_t$} & Input observations to the GrASPE model at time $t$  \\ 
\hline
\multirow{2}{4em}{$E_t$} & Ground truth label vector for a trajectory's success probability  \\ 
\hline
\multirow{2}{4em}{$\hat{E}_t$} & Predicted success probability vector from the GrASPE model  \\
\hline
\multirow{2}{4em}{$\mathcal{G}(\mathcal{V},\mathcal{W})$} & A connected graph represented by a set of vertices $\mathcal{V}$ and an adjacency matrix $\mathcal{W}$    \\ 
\hline
\multirow{1}{4em}{$r_{img}$} & Image reliability measure $\in [0,1]$  \\ 
\hline
\multirow{1}{4em}{$r_{point}$} & LiDAR point cloud reliability measure $\in [0,1]$ \\ 

\hline
\end{tabular}
\end{center}
\vspace{-17pt}
\label{tab:symbol_defn}
\end{table}

\begin{figure}[t]
      \centering
      \includegraphics[width=\columnwidth,height=4.3cm]{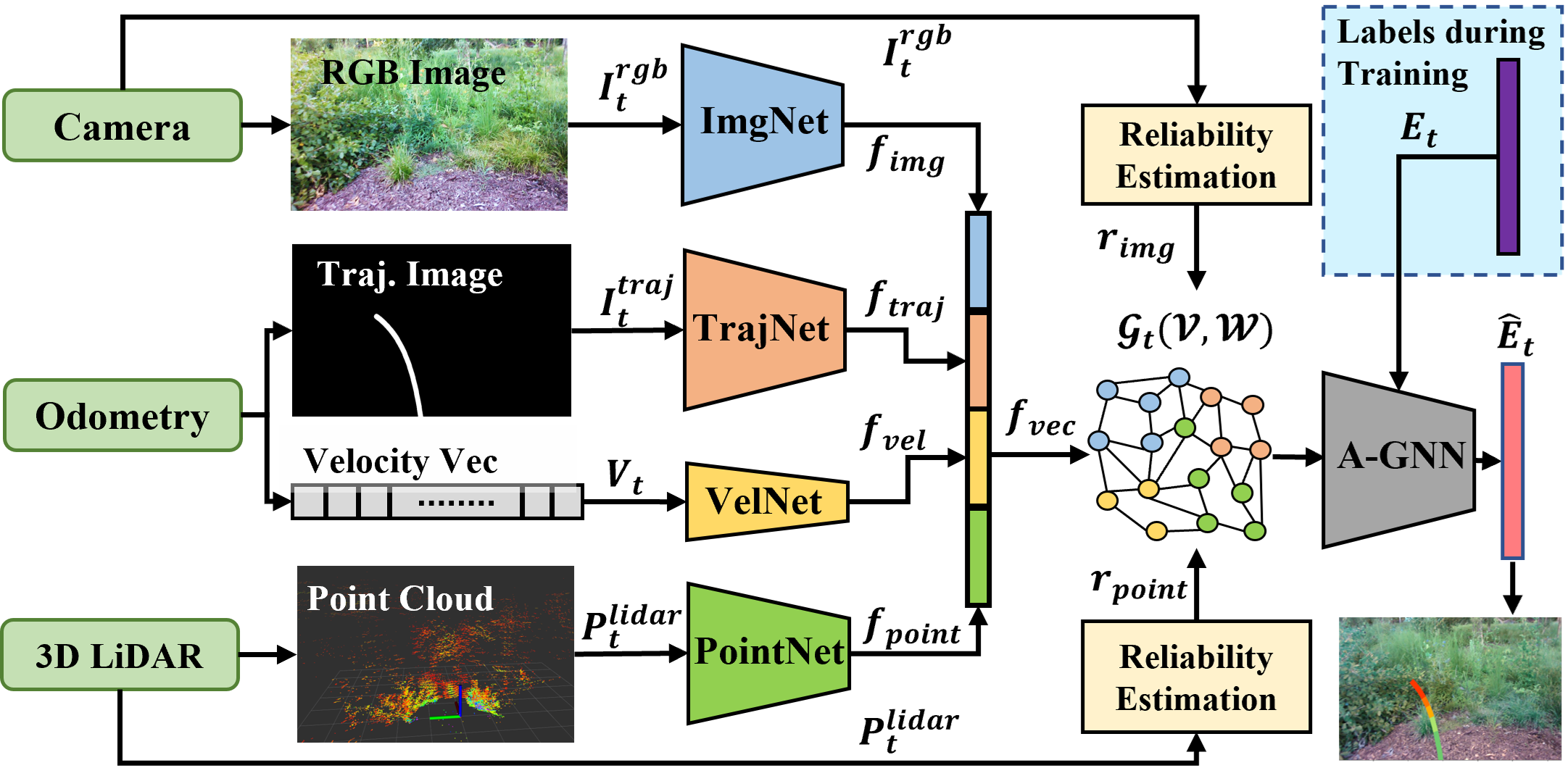}
      \caption {\small{GrASPE System Architecture for candidate trajectories' success probability estimation. We utilize RGB images, 3D point clouds, the robot's velocity history, and the predicted trajectory of the robot as an image to train a novel graph-based prediction model to estimate the given trajectory's probability of success in terms of navigation.  }  }
      \label{fig:sys_architecture}
      \vspace{-5pt}
\end{figure}

\invis{NO NEED TO HAVE A SEPARATE, SMALL BACKGROUND SECTION.
YOU SHOULD ALSO STATE THE PROBLEM CLEARLY!}

\subsection{Problem Formulation}

Our formulation for GrASPE can be stated as follows:
\begin{probform}
To predict the success probability vector $\hat{E}_t \in [0, 1]^T$ of a robot trajectory using its local multi-modal fused observations $O_t$ while accounting for the variable reliability ($r_{img}, r_{point}$) of the camera and LiDAR observations respectively. 
\end{probform}

We define \textit{success probability} as a trajectory's probability of not encountering collisions, stumbling, or similar modes of failure. The success probability is then used to evaluate trajectories that the robot can use for navigation. The overall system architecture of the prediction model is presented in Fig. \ref{fig:sys_architecture}.



\subsection{Graphs and Graph Neural Networks} \label{subsec:GNN}
Real-world data modalities (say images and point clouds) used for sensor fusion exhibit complex correlations between them. For instance, a rock captured in an image from a robot is represented by RGB values, as XYZ values in point clouds, and a spikes in the acceleration and angular velocity in a 6-DOF IMU when the robot runs over it. For accurate sensor fusion, these complex correlations cannot be represented by regular data structures (i.e. in the Euclidian domain). 

Representing the data streams as a graph offers the capability of modeling the complex correlations between them as weighted edge connections in the graph. Therefore, in our work, sensor data is processed and represented as a weighted graph $\mathcal{G} : = (\mathcal{V}, \mathcal{W})$, where $\mathcal{V} = { 1, 2, …, N}$ is a set of $N$ nodes/vertices associated with elements of the sensors' feature vectors. $\mathcal{W}$ is a weighted adjacency matrix with entries $\mathcal{W}_{i,j}$ representing the strength of the 
connection (edge) between feature elements $i$ and $j$.

Graph Neural Networks (GNNs) are a class of neural networks that operate on data represented as graphs. We incorporate Graph Convolution Network (GCN) \cite{gcn} and Graph Attention Network (GAT) \cite{agcn,rgat} operators in our GNN to encode and pay attention to the local graph structure and node embeddings. These networks are general forms of convolution and attention mechanisms and are capable of encoding node features and correlations between the graph nodes respectively. Especially, the GAT operator performs graph attention in four steps: Linear transformation, Leaky ReLU activation, Softmax normalization, and Multi-head attention which often outperforms traditional multi-head attention approaches on Euclidean data structures. 

\section{Our Algorithm: GrASPE} \label{sec:our-method}

\begin{figure}[t]
      \centering
      \includegraphics[width=\columnwidth,height=2cm]{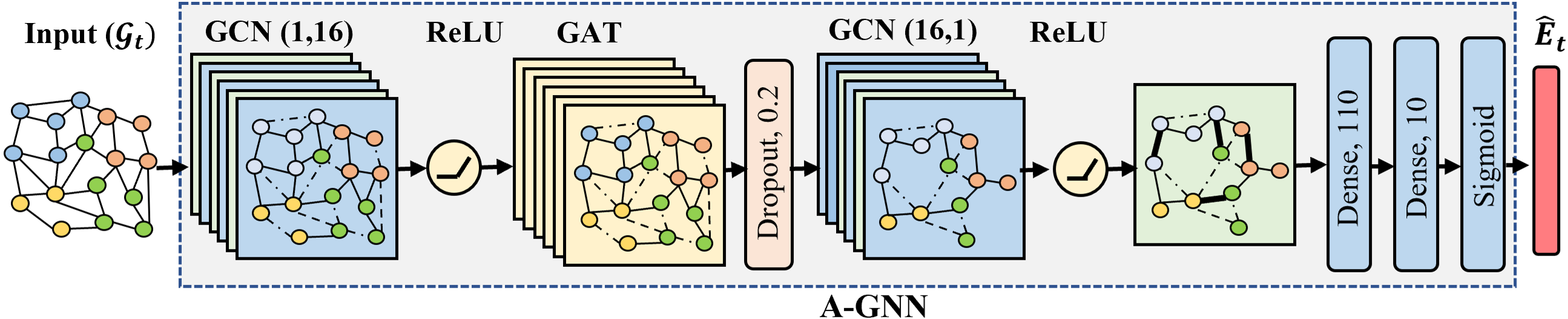}
      \caption {\small{Architecture of the attention-based graph neural network (A-GNN) used in our GrASPE system. The \textit{reliability-aware} feature graph $\mathcal{G}_t$ is fed into this A-GNN to encode and attend to the useful multi-modal sensory feature interactions.}  }
      \label{fig:agnn_architecture}
      \vspace{-15pt}
\end{figure}

In this section, we present the details of our proposed path evaluation algorithm. We first discuss the components of the prediction model trained for the trajectory's success probability estimation: 1. Acquiring multi-modal data and reducing its dimensionality; 2. Estimating camera and LiDAR's reliability;  3. Graph-based prediction model. Finally, we explain the details of how robust perception is used for planning.


\subsection{Multi-modal Observations}
We design our prediction model (i.e., success probability estimation model) using the following multi-sensor inputs as observations: RGB image $I_{t}^{rgb} \in \mathbb{R}^{w \times h \times 3}$, 3D point cloud $P_t^{lidar} \in \mathbb{R}^{3 \times N_{p}} $, robot's velocity history $V_t = \{(v,\omega)_{t-T+1},...,(v,\omega)_{t}\}$ and predicted trajectory from the robot's current velocity as an image $I_t^{traj} \in \mathbb{R}^{w \times h \times 1}$ (see Fig. \ref{fig:sys_architecture}). To obtain $I_t^{traj}$, we first extrapolate the robot's trajectory for the next $T = 10$ times steps based on the current velocity $(v,\omega)_t$. Then the resulting trajectory is projected to a blank image using homography projection. The RGB image and 3D point cloud capture visual and geometric features of the environment respectively. The robot's velocity history is used to account for the robot's recent behavior. The ground truth labels for predicting a trajectory's success are represented by a vector $E_t$ of 0's (for portions of the trajectory with failures) and 1's (for portions with successes) of length T.

Finally, the prediction model uses the observations $O_t = \big[I_{t}^{rgb}, P_t^{lidar},V_t,I_t^{traj}\big]$, and the ground truth labels to train to estimate the traversability of the predicted trajectory $\hat{E}_t = [p_t,p_{t+1},.., p_{t+T-1}]$ (i.e., success probability vector). Details of the observation data collection and ground truth labeling are described in Section \ref{subsec:dataset}.



\subsection{Multi-modal Feature Vector Generation} \label{sec:feature-vec-generation}

\invis{WHAT IS YOUR OVERALL GOAL HERE?}
Different sensor modalities capture different environmental features (images capture visual features, point clouds capture edges and surfaces), and fusing them effectively is non-trivial. Moreover, raw image and point cloud data processing are computationally expensive due to the high dimensionality. Therefore, we pre-process each sensor input using separate feature encoding networks to obtain dimension-reduced feature vector representations.

The sizes of these feature vectors are chosen empirically considering the trade-off between the resulting graph size (effects on the computation complexity) and the feature encoding quality (smaller feature vectors may not properly encode the input data).

\subsubsection{Visual Feature Extraction}
\label{subsec:feature_nets}
We utilize a ResNet\cite{resnet} based pipeline to extract image features $f_{img}$ from the camera RGB image $I_{t}^{rgb}$. We incorporate a CBAM \cite{woo2018cbam} module between the network layers to perform spatial and channel attention towards the important visual features. The input image $I_{t}^{rgb}$ is of size $320 \times 240$ (i.e., $w = 320$ and $h = 240$). The ImgNet branch in Fig. \ref{fig:sys_architecture} includes a ResNet18 backbone with CBAM layers added similar to \cite{woo2018cbam}. The output feature vector $f_{img}$ of dimension $40 \times 1$ is obtained by passing through a $Sigmoid()$ activation layer.

\subsubsection{LiDAR Feature Extraction} 
A Pointnet \cite{qi2017pointnet} based network is incorporated to extract point cloud features $f_{point}$ from the LiDAR data $P_t^{lidar}$. To reduce the data size, we restrict LiDAR point cloud to $[-\pi/2, \pi/2]$ field of view w.r.t. the robot's heading direction (see Fig. \ref{fig:point_reliability}). $P_t^{lidar}$ is of dimensions $3 \times N_p$, where $N_p = 10000$ is the number of points captured from the LiDAR at each time $t$. The output $f_{point}$ of dimensions $40 \times 1$ is obtained by passing through a $Sigmoid()$ activation layer at the end similar to ImgNet.

\subsubsection{Velocity Feature Extraction} 
A velocity vector $V_t$ of size $100 \times 1$ includes linear and angular velocities of the previous $50$ time steps (i.e. $v_t, \omega_t$ data of size $50 \times 2$ is reshaped to a vector $V_t$ of size $100 \times 1$). This $V_t$ is fed into 4 linear convolutional layers with dilation size $\{3,2,2,2\}$ in VelNet to obtain the velocity features $f_{vel}$ of length $20$. 

\subsubsection{Trajectory Feature Extraction} 
The predicted trajectory image $I_t^{traj} \in \mathbb{R}^{320 \times 240 \times 1}$ passes through 3 pooling and 3 convolutional layers with kernel size ${3,2,2}$, a flatten layer and $Sigmoid()$ activation layer to obtain the trajectory feature vector $f_{traj}$ of length $20$. Hence, the final feature vector $f_{vec}$ is of dimensions $120 \times 1$ after concatenating $f_{img}, f_{point}, f_{vel}$ and $f_{traj}$.
 
\subsection{Sensor Reliability Estimation} \label{subsec:reliability}
\invis{HOW IS RELIABILITY REPRESENTED (A SCALAR OR VECTOR)? WHY DO YOU CHOOSE SUCH}
RGB images and the point cloud data become unreliable in certain outdoor conditions (e.g. low luminance, scattering of laser rays, etc.). Even though the feature encoding networks proposed in the literature and in Section \ref{subsec:feature_nets} are capable of feature extraction from images and point clouds, they do not account for their reliability. 

\invis{DO NOT REFER TO TALL GRASS AGAIN. SOME OF THESE ISSUES SHOULD ONLY BE MENTIONED IN SECTION 1, NO NEED TO REPEAT THEM HERE. }

\invis{Hence, we consider the sensory inputs that contain sufficient features for navigation as reliable THIS SENTENCE IS UNCLEAR?. }
Therefore, we estimate the reliability of the image and point cloud inputs at each time step quantitatively using classical image and point processing methods that execute in real-time \cite{fast2,zhang2014loam}. We further assume that the instantaneous robot velocities obtained from its odometry are not significantly affected by the environment. Therefore, it is considered reliable during operation. 

\subsubsection{Image Reliability Estimation} 
\begin{figure}[t]
    \centering
    \includegraphics[width=\columnwidth,height=6cm]{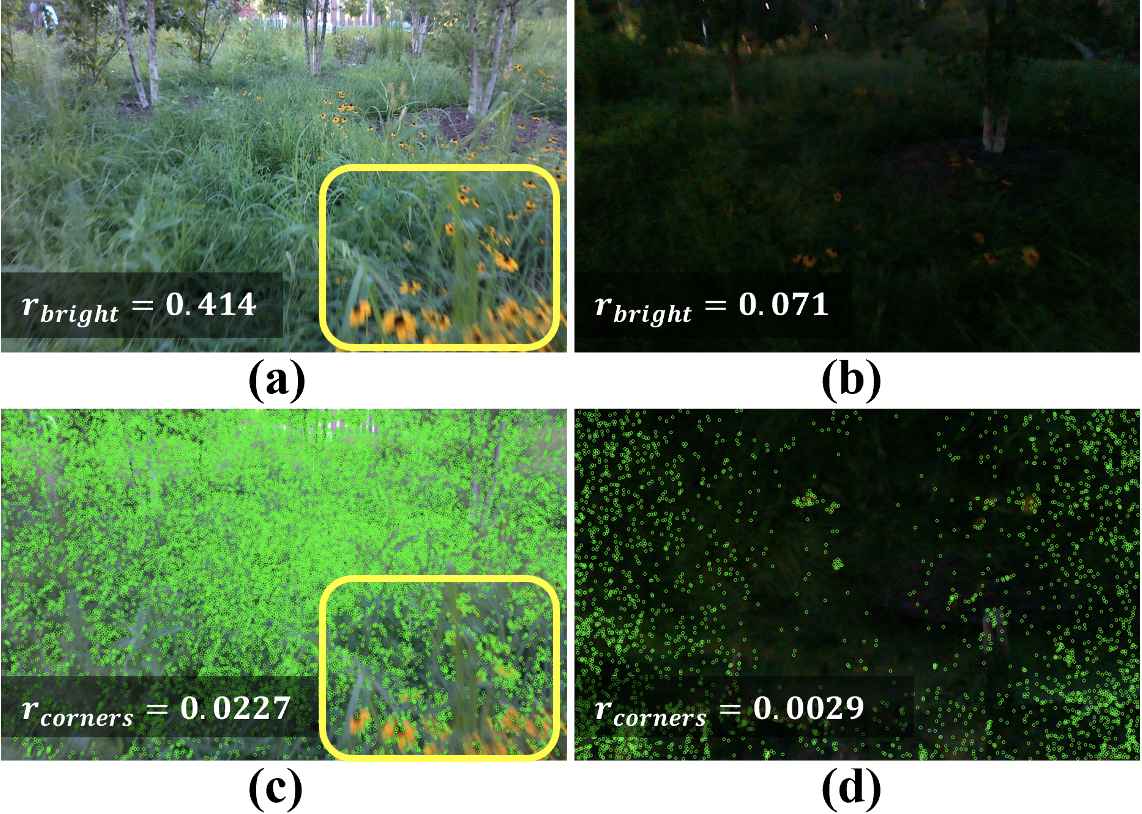}
    \caption{\small{\textbf{Image Reliability Estimation:} We estimate input image reliability based on two factors: 1. Overall lighting condition 2. Feature richness. (a) and (b) demonstrate images of the same scene under two different lighting conditions. (c) and (d) presents the FAST \cite{fast,fast2} features extracted from the images (a) and (b). We observe that the regions in images with poor lighting and motion blur (yellow rectangle in (a) and (c)) have a significantly lower number of features. Similarly, the motion blur and camera occlusions result in images with low reliability measures. }}
    \label{fig:img_reliability}
\end{figure}

We consider that the images captured by the camera have high reliability if they have: 1. High image brightness, and 2. Availability of visual features such as corners, edges, etc \cite{mojsilovic2002isee}. See Fig. \ref{fig:img_reliability} and \ref{fig:point_reliability} for sample reliability comparisons. To estimate the image brightness, we first convert the input RGB image $I_{t}^{rgb}$ to a gray-scale image $I_{t}^{gray}$. Then we calculate the Root Mean Square(RMS) value of the histogram distribution of $I_{t}^{gray}$ as follows to obtain brightness estimation,

\invis{IS THIS A STANDARD METHOD? IF SO, PROVIDE A REFERENCE?}
\vspace{-10pt}
\begin{equation}
  r_{bright} = RMS( hist(I_{t}^{gray}))
  \vspace{-4pt}
\end{equation}

Here, $hist$ is the histogram operator and $RMS$ is the root mean square operator where $RMS(x) = \sqrt{\frac{1}{n}\sum_{i=1}^n x_{i}^2}$.

To estimate the availability of useful features in the image, we calculate the number of corner features $n_{c}$ in the input image $I_{t}^{rgb}$ using FAST (Features from Accelerated Segment Test) algorithm \cite{fast, fast2}. These corner features are fast to compute and inherently reflect the non-blurriness and well-lit condition, therefore a good measure of the image's reliability. We consider the image is feature rich if the $n_c$ is higher than a threshold. Hence, $r_{corners} = \frac{n_c}{(w \times h)}$. Here, $w$ and $h$ are the height and width of the input RGB image.
The final image reliability measure is obtained as a normalized scalar value $r_{img} \in [0,1]$ using the weighted sum of $r_{bright}$ and $r_{corners}$ : 

\invis{WHY DO YOU USE THIS FORMULA}
\vspace{-5pt}
\begin{equation}
  r_{img} = \alpha_b r_{bright} + \alpha_c r_{corners} 
  \vspace{-2pt}
\end{equation} 

\subsubsection{Point Cloud Reliability}

We observe that the lidar point cloud is heavily distorted in the presence of unstructured vegetation leading to poor estimation of the surrounding objects' geometries. Therefore, to evaluate the point cloud reliability, we perform edge and planar feature extraction to calculate the number of 3D features available in the point cloud $P_t^{lidar}$ at a given time $t$. High numbers of edges and planes denote low scattering (i.e., structured) in the point cloud.

Let $X_l$ be the $l^{th}$ 3D point in the input point cloud $P_t^{lidar}$ and let $\mathcal{M}$ be the set of points in the neighborhood of $X_l$ acquired from an instance of the point cloud. Then, we can obtain the local surface smoothness evaluation factor $c_l$ of the point $l$ as,

\vspace{-4pt}
\begin{equation}
\label{eq:point_feature}
  c_l = \frac{1}{||X_l||.|\mathcal{M}|} \Big\lVert \sum_{k \in \mathcal{M}, k \neq l} \big( X_l - X_k \big) \Big\rVert,
  \vspace{-3pt}
\end{equation}
where $ X_k$ with $k = 1,2,..,|\mathcal{M}| $ are the coordinates of the 3D points in the set $\mathcal{M}$. Zhang et al. \cite{zhang2014loam} demonstrate that the points with higher and lower $c_l$ values obtained from equation \ref{eq:point_feature} belong to edge (non-smooth) and planar (smooth) features respectively. Hence, we define two threshold values $c_{max}$ and $c_{min}$ as the minimum and maximum smoothness thresholds to consider a given point on a surface as an edge or a plane respectively. The resulting edge and planar feature point sets can be denoted as: 

$\mathcal{S}_{edge} = \{l | c_l \geq c_{max}, l \in P_t^{lidar} \} $ and $\mathcal{S}_{planar} = \{l | c_l \leq c_{min}, l \in P_t^{lidar} \}$ respectively. 

\begin{figure}[t]
    \centering
    \includegraphics[width=\columnwidth,height=3cm]{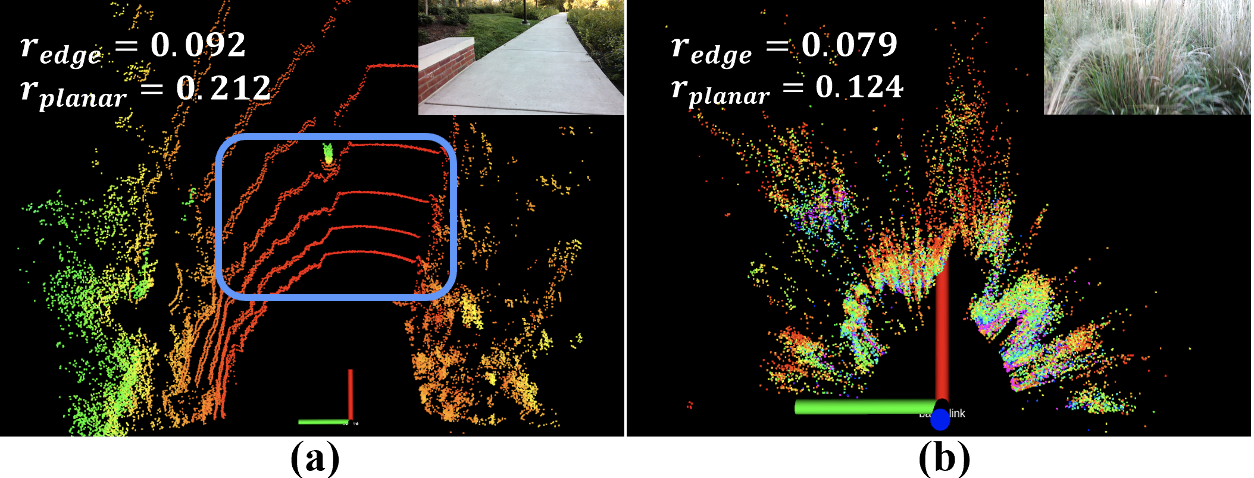}
    \caption{\small{\textbf{Point cloud Reliability Estimation:} We estimate point cloud reliability based on the edge and planar feature availability. (a) Top view of a point cloud observation with many edge and planar features (blue rectangle) and the corresponding RGB image (top right) (b) Top view of a distorted point cloud observed in tall grass environments. Point clouds are reliable in open and structured environments and become unreliable in cluttered vegetation and adverse weather.}}
    \label{fig:point_reliability}
\end{figure}
 
\vspace{-4pt}
\begin{equation}
  r_{edge} = \frac{|\mathcal{S}_{edge}|}{|P_t^{lidar}|}, \quad r_{planar} = \frac{|\mathcal{S}_{planar}|}{|P_t^{lidar}|}.
  \vspace{-4pt}
\end{equation}
Finally, the point cloud reliability measure $r_{point}$ is derived as a weighted combination of 
$r_{edge}$ and $r_{planar}$ using two tunable parameters $\beta_e$ and $\beta_p$.
\vspace{-5pt}
\begin{equation}
  r_{point} = \beta_e r_{edge} + \beta_p r_{planar}.
  \vspace{-5pt}
  \label{eq:pointcloud_reliability}
\end{equation}
 Fig. \ref{fig:point_reliability}(c) and (d) presents an example comparison of $r_{edge}$ and $r_{planar}$ in real outdoor scenarios.

\subsection{Reliability-aware Graph Construction} \label{subsec:gnn-method}

Once the sensor inputs are encoded into feature vectors (Section \ref{sec:feature-vec-generation}), we concatenate them to obtain a combined vector $f_{vec}= [f_{img}, f_{point}, f_{traj}, f_{vel}]$. The $i^{th}$ element of $f_{vec} \in [0,1]^N$ ($N = 120$) becomes the $i^{th}$ node of the graph representation for fusion.



The weighted adjacency matrix $\mathcal{W}$ represents the strength between the graph nodes (i.e. edge weights). i.e., the connectivity between the encoded feature elements in $f_{vec}$. Typically, the edge weights are calculated using a distance measure (absolute difference or L2 norm). However, we modify these edge weights (especially between the nodes corresponding to the image and point cloud features) based on the reliability measures (Section \ref{subsec:reliability}) as follows. 


Let $\mathcal{W}_{i,j}$ be the edge weight between the feature elements $i$ and $j$ in $f_{vec}$. 
\vspace{-5pt}
\begin{equation}
\mathcal{W}_{i,j} = \begin{cases}
    exp^{-\lbrace \lambda \cdot |f_{vec}(i)-f_{vec}(j)| \cdot (2 -r_{i,j})\rbrace},& \text{if } i\neq j\\
    0,              & \text{otherwise},
\end{cases}
\vspace{-5pt}
\label{eq:graph_weights}
\end{equation}
where, $\lambda > 0$ is a tunable scalar and $r_{i,j} \in [0,1]$ represents the reliability measure between the node $i$ and $j$ as follows,
\vspace{-5pt}
\begin{equation}
r_{i,j} = \begin{cases}
    r_{img},& \text{if } \{i,j | i \in f_{img} , j \notin f_{point} \} \\
    r_{point},& \text{if } \{i,j | i \notin f_{img} , j \in f_{point} \} \\
    \frac{1}{2} \big( r_{point}+r_{img} \big) ,& \text{if } \{i,j | i \in f_{img} , j \in f_{point} \} \\
    1,              & \text{otherwise}. \\
    
\end{cases}
\end{equation}

Hence, we refer to the derived graph representation $\mathcal{G}_t$ of the feature vector $f_{vec}$ as a \textit{reliability-aware} feature graph.

\subsection{A-GNN Architecture} \label{sec:a-gnn-arch}
We design a light-weight Attention-based Graph Neural Network (A-GNN) presented in Fig.\ref{fig:agnn_architecture} to predict the success probability vector $\hat{E}_t$ for a given trajectory. We first pass our reliability-aware feature graph $\mathcal{G}_t$ through a 16-channel graph convolutional layer (GCN) to encode the node features. Secondly, a graph attention network (GAT) is utilized to identify the important neighbors (i.e., strongly correlating neighbors) of each node of the 16-channel graphs. Next, a dropout layer is incorporated for regularization to minimize the risk of network over-fitting. The output graph is obtained after passing through another GCN layer. We use the $ReLU$ activation layer after each GCN layer. The output graph is concatenated into a vector and passed through two dense layers and $sigmoid$ activation to obtain the output prediction vector $\hat{E}_t$. Hence, the overall prediction pipeline can be treated as a mapping function $\psi_{GrASPE}: O_t \rightarrow \hat{E}_t$.

\subsection{Theoretical Validation of Graph Construction}
\begin{lemma} 
GrASPE's reliability aware feature graph $\mathcal{G}_t$ generated using the weight matrix $\mathcal{W}$ ensures that $\mathcal{G}_t$ is an undirected graph with non-negative weights. 
\label{lem:undirected_graph}
\end{lemma}

\begin{proof}
Let $w_{i,j} \in \mathcal{W}$. By construction, $r_{i,j} = r_{j,i} \, \forall \, i,j, \text{and} \, i \neq j $ $\implies w_{i,j} = w_{j,i} \, \forall \, i,j, \text{and} \, i \neq j$ and $w_{i,j} = 0 \, \forall \,  i,j, \text{and} \, i = j.$ This implies that $\mathcal{W}$ is symmetric and $\mathcal{G}_t$ is undirected.\\

\no Let $m_{i,j }= \lambda |f_{vec}(i)-f_{vec}(j)| r_{i,j}$ from the Eq.\ref{eq:graph_weights}. Then, $m_{i,j} \geq 0, \, \forall \, i,j \,$ because $\lambda > 0$ and  $r_{i,j} \in [0,1]$. Therefore, $exp^{-m_{i,j}} > 0 \, \forall \, i,j, i \neq j \implies w_{i,j} \geq 0 \, \forall \, i,j \implies \mathcal{W}$ has non-negative weights.
\end{proof}

GCN layers in our A-GNN perform spectral graph convolution which requires the input graph's Laplacian $\mathcal{L}$ to be spectrally decomposable \cite{gcn}    (i.e., eigen decomposition on real symmetric matrices). We prove that our graph construction leads to a spectrally decomposable graph Laplacian matrix using the Lemma \ref{lem:laplacian}. 

From Graph theory, we consider the graph Laplacian $\mathcal{L} = \mathcal{D} -\mathcal{W}$, where $\mathcal{D}$ is a diagonal matrix with $\mathcal{D}_{i,i} = deg(v_i)$. Here, $deg(v_i)$ is the degree of a vertex which is a measure of the number of edges terminating at that vertex. In this context, we consider $\mathcal{D}_{i,i} = \sum_j w_{i,j}$. By construction and from Lemma \ref{lem:undirected_graph}, we can observe that $\mathcal{D}$ is real and $\mathcal{W}$ is hermitian (i.e., $w_{i,j} = w_{j,i}^*$).

\begin{lemma} 
Laplacian matrix $\mathcal{L}$ of the graph $\mathcal{G}_t$ is spectrally decomposable. 
\label{lem:laplacian}
\end{lemma}

\begin{proof}
From Lemma \ref{lem:undirected_graph}, $ Re(w_{i,j}) \geq 0 \, \forall \, i,j$. Consider a real valued $x$,\\
\begin{equation}
 \begin{aligned} 
 x^T\mathcal{L}x 
 &= \sum_{i,j} w_{i,j} (x_i-x_j)^2 \quad \because \mathcal{L} =\mathcal{D} -\mathcal{W}\\
 &= \sum_{i < j} w_{i,j} (x_i-x_j)^2 +\sum_{i>j} w_{i,j} (x_i-x_j)^2\\
 &= \sum_{i < j} w_{i,j} (x_i-x_j)^2 +\sum_{i>j} w_{j,i}^* (x_i-x_j)^2\\
&= \sum_{i < j} w_{i,j} (x_i-x_j)^2 +\sum_{i<j} w_{i,j}^* (x_j-x_i)^2\\
&= \sum_{i < j} w_{i,j} (x_i-x_j)^2 +\sum_{i<j} w_{i,j}^* (x_i-x_j)^2\\
&= \sum_{i < j} (w_{i,j} + w_{i,j}^*)(x_i-x_j)^2 \\
&= 2\sum_{i < j}Re (w_{i,j})(x_i-x_j)^2  \geq 0 \\
&\implies \mathcal{L} \quad \textrm{is Positive semi-definite.}
\end{aligned}
\vspace{-2pt}
\end{equation}
$\therefore$  Symmetric matrix $\mathcal{L}$ has all real eigenvalues. Further, the corresponding eigenvectors $u_1,..,u_N$ can be taken to be orthonormal by,
\vspace{-5pt}
\begin{equation}
u_i^T u_j = \begin{cases}
    1,& \text{if } i = j\\
    0,    & \text{if } i\neq j,
\end{cases}
\vspace{-2pt}
\end{equation}

\no Therefore, from the Spectral theorem \cite{spectral_theorem_Hall}, $\mathcal{L}$ can be spectrally decomposed as, $\mathcal{L} = U \Lambda U^T$, where 
$U$ is the eigenvector matrix, and $\Lambda$ denotes the diagonal matrix of sorted eigenvalues.

\end{proof}

\begin{algorithm}[t]
	\begin{algorithmic}[1]
        \STATE \textbf{Input}: $goal, obs, d_{goal},  I_{t}^{rgb}, P_t^{lidar},V_t$
        \STATE \textbf{Output}: $(v_{(t+1)},\omega_{(t+1)})$
		\STATE \textbf{Initialize} :   $T,\Delta t, v_{max},\omega_{max}, \dot{v}_{max},\dot{\omega}_{max},\alpha_b,\alpha_c,\beta_e,\beta_p,$ \\ $\gamma_1,\gamma_2,\gamma_3,r_{th}, d_{th}$ 
        \STATE $V_s = \{(v,\omega) | v \in [0,v_{max}], \omega \in [-\omega_{max},\omega_{max}]\}$ \\
        \WHILE{$d_{goal} \geq d_{th}$}
        
        \STATE $V_d = \{(v,\omega) | v \in [v-\dot{v}_{max}\Delta t,v+\dot{v}_{max}\Delta t], \omega \in [\omega-\dot{\omega}_{max}\Delta t,\omega+\dot{\omega}_{max}\Delta t]\}$
        \STATE $r_{point} = PointcloudRelibaility(P_t^{lidar})$ using the Eq.\ref{eq:pointcloud_reliability}.
        \IF{$r_{point} \leq r_{th}$}
        \STATE $V_r = V_s \cap V_d $
        \STATE $Q(v,\omega) = \sigma\big(\gamma_1 . heading(v,\omega) + \gamma_3 . vel(v,\omega) \big)$ 
        \ELSE
        \STATE $V_a = ObstacleFree(obs,V_s)$
        \STATE $V_r = V_s \cap V_d \cap V_a$ 
        \STATE $Q(v,\omega) = \sigma\big(\gamma_1 . heading(v,\omega) + \gamma_2 . dist(v,\omega) + \gamma_3 . vel(v,\omega) \big)$ 
        \ENDIF
        \STATE $(v_t^*,\omega_t^*) = argmax(Q(v,\omega))$
        \STATE $I_t^{traj} = GenerateTrajectoryImage(v_t^*,\omega_t^*,T)$
        \STATE $O_t = \big[I_{t}^{rgb}, P_t^{lidar},V_t,I_t^{traj}\big]$ 
        \STATE $\hat{E}_t = \psi_{GrASPE}(O_t)$
        \IF{$min(\hat{E}_t) \geq e_{th}$}
        \RETURN$(v_t^*,\omega_t^*)$
        \ELSE   
        \STATE $V_r = V_r \setminus (v_t^*,\omega_t^*)$
        \STATE \textbf{Goto to step 8}
        \ENDIF
        \ENDWHILE
	\end{algorithmic}
	\invis{\caption{Efficient Navigation with Adpative Heavy-tailed Reinforce (HTRON)}}
	\caption{GrASPE based Outdoor Navigation}
	\label{algo:one}
\end{algorithm}

Hence, having an undirected graph with non-negative weights and a spectrally decomposable Laplacian enables us to perform the graph convolution operation on our feature graph $\mathcal{G}_t$ to encode and learn its complex node embedding (i.e., the correlation between adjacent feature nodes) significantly better compared to CNNs and other architectures (see Benefits of A-GNN in Section \ref{sec:results}).


\subsection{Reliability-aware Planning} \label{subsec:dwa}
Our local planner adapts the Dynamic Window Approach (DWA)~\cite{fox1997dwa} to perform navigation while evaluating the success probabilities of the generated trajectories using GrASPE. Our overall algorithm is presented in Algorithm \ref{algo:one}. We explain the algorithm and the symbols used here. 

$V_s$ (line 4) is defined as the space of all the possible linear and angular robot velocities $(v, \omega)$. While the robot's distance to its goal ($d_{goal}$) is greater than a threshold (line 5), DWA computes $V_a$, the set of collision-free velocities, and $V_d$, the set of reachable velocities based on the robot's acceleration constraints (lines 6 and 7). Next, the reliability of the point cloud $r_{point}$ obtained from the robot is calculated and compared against a threshold (lines 8 and 9). If the point cloud is unreliable, implying that the detected obstacles and the calculated collision-free set $V_a$ are erroneous, the search space $V_r$ for choosing the robot's velocity is restricted to the intersection of $V_s$ and $V_d$.

DWA chooses the velocity belonging to $V_r$ that maximizes an objective function $Q(v, \omega)$ as the final velocity for navigation. $Q(v, \omega)$ consists of three terms: $heading(.), dist(.)$ and $vel(.)$ which quantify the robot's heading towards the goal, distance to the closest obstacle in the trajectory, and the forward velocity of the robot, respectively.

If the point cloud is unreliable, the $dist(.)$ function in $Q(v, \omega)$ is unreliable. Therefore, we omit it from the objective function (line 11). Otherwise, DWA's original search space and objective function are used (lines 13 and 14) to compute the optimal/maximizing velocities $(v_t^*,\omega_t^*)$ (line 16). Lines 17-19 use GrASPE to evaluate the optimal trajectory's success probability. If the optimal trajectory has a high success probability, the robot uses it for navigation (lines 20-21). Otherwise, it is removed from the search space and a new optimal velocity is computed (lines 22-24). $V_r$ typically has around 12 to 16 $(v, \omega)$ pairs making the search tractable.

\begin{figure*}[t]
    \centering
    \includegraphics[width=2\columnwidth,height=6.8cm]{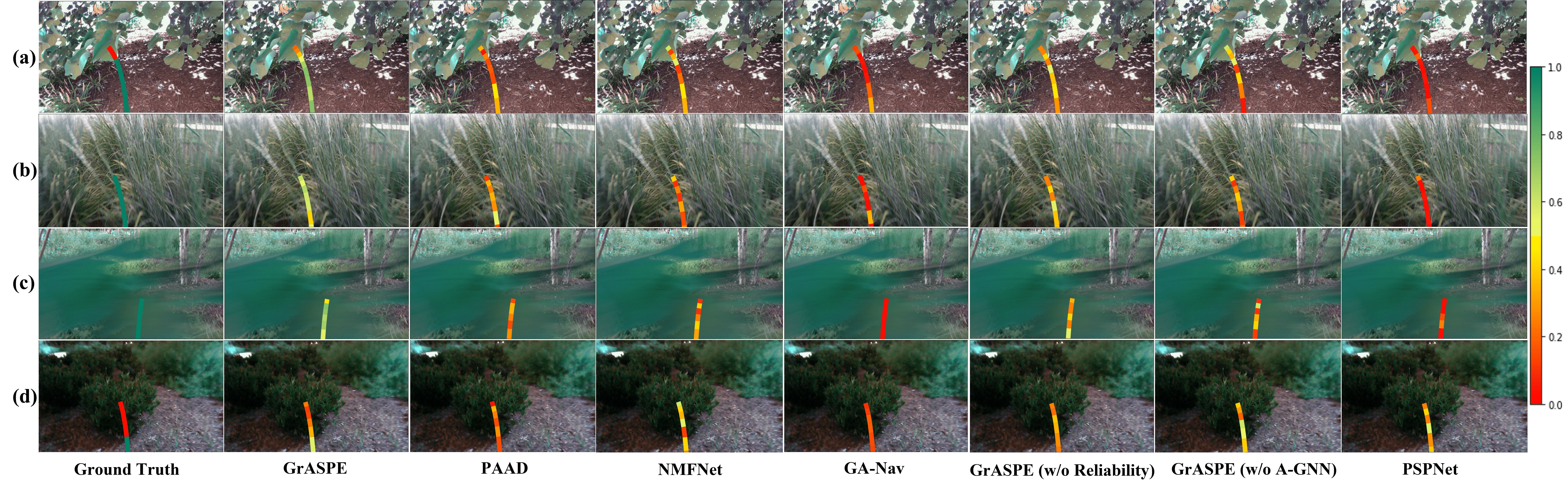}
    \caption{\small{Navigability predictions from GrASPE(Ours), PAAD \cite{ji2022proactive_ad}, NMFNet \cite{nguyen2020nmfnet}, GA-Nav \cite{ganav}, DWA \cite{fox1997dwa}, GrASPE w/o reliability, GrASPE w/o A-GNN and PSPNet \cite{pspnet} under different environmental conditions compared to the Ground Truth. (a) Hanging leaves; (b) Pliable tall grass; (c) Camera occlusion; (d) Low-light condition. The gradient on the right denotes the success probabilities. GrASPE predictions outperform the other methods under varying conditions that are critical for robot navigation in unstructured outdoor environments. Our method results in a 13-15\% decrease in the false positive rate during navigation.}}
    \label{fig:prediction_fig}
    \vspace{-5pt}
\end{figure*}


\section{Results and Analysis} \label{sec:results}

We detail our method's implementation and experiments conducted on a Spot robot. Then, we perform ablation studies and comparisons to highlight the benefits of our algorithm.

\subsection{Implementation}
GrASPE is implemented using PyTorch and PyTorch Geometric (PyG). The prediction model is trained in a workstation with an Intel Xeon 3.6 GHz processor and an Nvidia Titan GPU using real-world data collected from a Spot robot. The robot is equipped with an Intel NUC 11 (a mini-PC with Intel i7 CPU and NVIDIA RTX 2060 GPU), a Velodyne VLP16 LiDAR, and an Intel RealSense L515 camera.

\subsection{Dataset}
\label{subsec:dataset}
The multi-modal dataset used in this work is collected by operating the Spot robot under different lighting conditions in an outdoor field that includes bushes, small trees, hanging leaves, and grass regions of different heights and densities.   We incorporate a randomized planner to collect $I_{t}^{rgb}$, $P_t^{lidar}$ and $V_t$ from the robot to minimize human effort. The ground truth labels are created part manually and part automatically. The non-traversable and traversable portions in a trajectory are marked with 0's and 1's manually. Trajectories that are completely traversable during data collection are marked automatically with 1's. The final dataset includes a set of observations and label pairs $(O_t,E_t)$ for each time step $t$. 
The training set includes 28721 positive samples and 8463 negative samples whereas the test set contains 4537 positive samples and 1096 negative samples. 


\begin{figure*}[t]
    \centering
    \includegraphics[width=17cm,height=3.7cm]{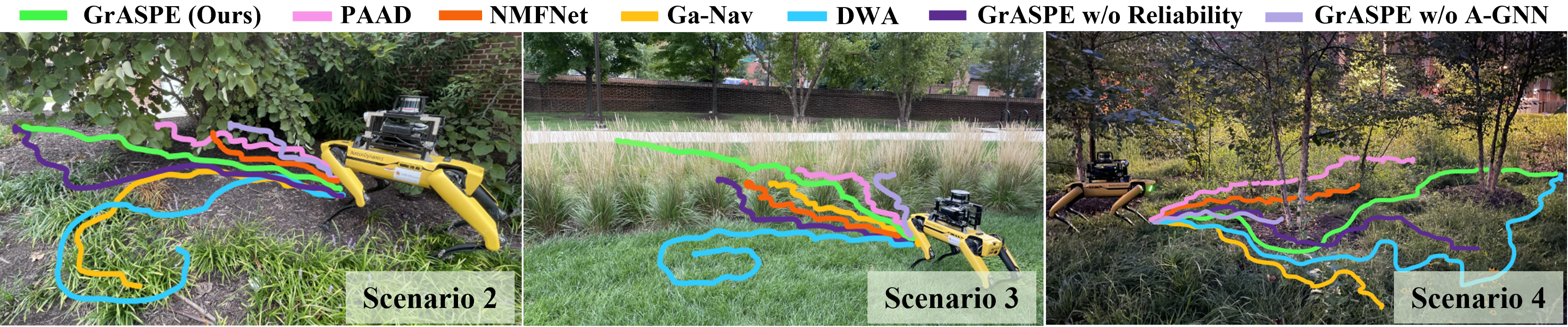}
    \caption{\small{Spot robot navigation in challenging outdoor scenarios using GrASPE (ours), PAAD \cite{ji2022proactive_ad}, NMFNet \cite{nguyen2020nmfnet}, GA-Nav \cite{ganav}, DWA \cite{fox1997dwa}, GrASPE w/o reliability, and GrASPE w/o A-GNN. The test environments are different from the training scenarios. GrASPE is able to successfully continue navigation in varying environmental conditions.}}
    \label{fig:navigation_comparisons}
    \vspace{-15pt}
\end{figure*}

\subsection{Evaluations} 
We use the following evaluation metrics to compare our method's performance against: PAAD \cite{ji2022proactive_ad}, NMFNet \cite{nguyen2020nmfnet}, GA-Nav \cite{ganav}, and DWA \cite{fox1997dwa} algorithms. 

PAAD is a proactive anomaly detection-based navigation method that fuses images, 2D LiDAR scans, and the robot's future trajectory using Multi-head attention. 
NMFNet is a deep multi-modal fusion network that has three branches: 2D laser, RGB images, and point cloud data for action prediction.
GA-Nav is a multi-modal navigation framework that combines image semantic segmentation with elevation maps for traversability estimation. 
 DWA is a local planner that utilizes a 2D LiDAR scan for obstacle avoidance. 

We further compare the GrASPE model without reliability estimations (GrASPE w/o reliability) and without the A-GNN model (GrASPE w/o A-GNN) for ablation studies. 
PAAD, NMFNet, and the two ablation models are trained using the same dataset (Section \ref{subsec:dataset}) for evaluations. GA-Nav is trained on publicly available outdoor datasets (RUGD \cite{RUGD2019IROS} and RELLIS-3D \cite{jiang2020rellis3d}) since it requires pixel-level segmentation labels.


\no \textbf{Success Rate} - The number of successful goal-reaching attempts (while avoiding non-pliable vegetation and collisions) over the total number of trials.

\no \textbf{Normalized Trajectory Length} - The ratio between the robot's trajectory length and the straight-line distance to the goal in both successful and unsuccessful trajectories. 

\no \textbf{False Positive Rate (FPR)} - The ratio between the number of false positive predictions (i.e., actually non-traversable trajectories predicted as traversable) and the total number of actual negative (non-traversable) samples encountered during a trial. We report the average over all the trials. 

\no \textbf{False Negative Rate  (FNR)} - The ratio between the number of false negative predictions (i.e., actually traversable trajectories as non-traversable) and the total number of actual positive samples encountered during a trial. We report the average over all the trials. 

\subsection{Test Scenarios}
We evaluate our navigation method's performance in the following four outdoor test scenarios that differ from the training environments. At least 10 trials are conducted in each scenario.

\begin{itemize}
\item \textbf{Scenario 1} - Contains trees, pliable and non-pliable plants and bushes (Fig. \ref{fig:cover_image}).

\item \textbf{Scenario 2} - Contains hanging leaves and trees where the sensors can undergo partial and complete occlusions (Fig. \ref{fig:navigation_comparisons}(a)).

\item \textbf{Scenario 3} - Contains open ground and pliable tall grass regions (Fig. \ref{fig:navigation_comparisons}(b)).

\item \textbf{Scenario 4} - Contains trees and vegetation under low light conditions (Fig. \ref{fig:navigation_comparisons}(c)).
\end{itemize}

\subsection{Analysis and Discussion}
The qualitative and quantitative results are presented in Fig. \ref{fig:prediction_fig}, \ref{fig:navigation_comparisons}, and Table \ref{tab:comparison_table} respectively. Qualitative results in Fig. \ref{fig:prediction_fig} demonstrate that our method is able to differentiate trajectories with high and low success probabilities even during camera occlusions and low light conditions (see Fig. \ref{fig:prediction_fig}(c), (d)). GrASPE also provides accurate success prediction in cluttered regions in Fig. \ref{fig:prediction_fig}(a), (b) where point cloud encounters heavy distortions. 

We further observe that GrASPE outperforms all the comparison methods in terms of success rate, false positive rate, and false negative rate when navigating in highly unstructured outdoor regions. This is primarily due to the improved trajectory success probability predictions generated by our algorithm using \textit{reliability-aware} sensor fusion. 

PAAD and NMFNet demonstrate comparative perception and navigation performance in moderately complex environments since they are designed for multi-modal fusion. Especially, PAAD is capable of predicting potential navigation failures under slightly distorted sensor observations which result in reasonably lower FPR and FNR for Scenarios 2 and 3. However, PAAD fails to continue with successful navigation when the robot enters to extremly challenging regions such as cluttered vegetation (i.e., tall grass in Scenario 3) and hanging leaves in Scenario 2. NMFNet shows severe navigation performance degradation in all scenarios due to lack of potential failure prediction capabilities (i.e., only performs reactive prediction based on current observation).

 2D LiDAR based methods such as DWA perform well in less cluttered regions such as scenario 4. However, thin, cluttered, and pliable vegetation are also detected as obstacles by DWA. Hence, it could not succeed in Scenarios 1 and 3 even once during testing. Instead, the robot gets stuck near the cluttered region and attempts to find free space by rotating. In scenario 4, DWA generates longer trajectories to avoid thin pliable grass regions while GrASPE traverses through them to generate relatively shorter trajectories. Further, PAAD, NMFNet and GA-Nav result in incomplete trajectories mostly under visually challenging scenarios 2 and 4, which leads to normalized trajectory length values of less than 1. We report them since they measure the progress toward the goal. 
 
GA-Nav uses a point cloud based elevation map along with RGB images to estimate terrain traversability. We observe that the distorted point clouds generate misleading estimations of terrain elevation. 
This results in a significantly lower success rate for GA-Nav in scenario 2. However, GrASPE-based navigation is robust even when the robot's camera is 100\% occluded, and the LiDAR is partially occluded. It uses the partially reliable LiDAR and robot's odometry to avoid collisions and navigate to its goal. 

\textbf{Failure cases:} There are trials in scenarios 1 and 2 where both the camera and LiDAR are highly occluded and unreliable. In such cases, GrASPE-based navigation does not have any feedback to avoid obstacles and non-traversable regions and could cause collisions or get stuck. We note, however, that our method still significantly outperforms the most recent and related works such as PAAD and NMFNet. 

 \textbf{Benefits of Attention-based GNN (A-GNN) :} In this ablation, we feed $f_{vec}$ to a set of 1D convolution layers, instead of using the graph representation and our A-GNN. We observe this GrASPE w/o A-GNN pipeline cannot outperform our GrASPE system qualitatively or quantitatively after training and testing with our dataset. This is primarily due to the strong feature correlation learning capabilities of our A-GNN.

\textbf{Ablation Study on Reliability Measures:} We compare the navigation and prediction performance of GrASPE with and without using the reliability terms in equation \ref{eq:graph_weights}. GrASPE without reliability leads to erroneous predictions when one or more sensory inputs are heavily distorted. Hence, even though the GrASPE model without reliability can navigate reasonably well compared to other methods, it still could not complete the navigation tasks consistently which result in a lower success rate and higher false positive and negative rates. For example in Scenario 3 in Fig. \ref{fig:navigation_comparisons}, the robot navigates successfully until the point cloud and camera are distorted in the tall vegetation region.

\textbf{Inference Rate:} Table \ref{tab:inference_table} compares GrASPE's inference rate with the other methods which use RGB images and LiDAR scan or point clouds. The rates are calculated from the instant sensor data is obtained to when a velocity is calculated. We observe that GrASPE has a low inference rate due to the computationally heavy GNN backbone. However, the design choices in our GNN architecture (Section \ref{sec:a-gnn-arch}) such as dropout and GAT layers (instead of deep GCN backbones that are computationally expensive)
help keep the rates high enough for navigation. To the best of our knowledge, our GNN architecture is the first to achieve real-time execution on a robot mounted computer.  

\begin{table}
\resizebox{\columnwidth}{!}{%
\begin{tabular}{ |c |c |c |c |c |c |} 
\hline
\textbf{Metrics} & \textbf{Method} & \multicolumn{1}{|p{1cm}|}{\centering \textbf{Scenario} \\ \textbf{1}} & \multicolumn{1}{|p{1cm}|}{\centering \textbf{Scenario} \\ \textbf{2}} & \multicolumn{1}{|p{1cm}|}{\centering \textbf{Scenario} \\ \textbf{3}} & \multicolumn{1}{|p{1cm}|}{\centering \textbf{Scenario} \\ \textbf{4}}\\ [0.5ex] 
\hline
\multirow{7}{*}{\rotatebox[origin=c]{0}{\makecell{\textbf{Success}\\\textbf{Rate (\%)}}}} 
 & DWA \cite{fox1997dwa} & 0 & 20 & 0 & 70   \\
 & GA-Nav \cite{ganav} & 10 & 10 & 30 & 10 \\
  & NMFNet \cite{nguyen2020nmfnet}  & 30 & 10 & 40 & 20 \\
 & PAAD \cite{ji2022proactive_ad}  & 40 & 40 & 50 & 40 \\
 & GrASPE without A-GNN & 10 & 20 & 10 & 0 \\
 & GrASPE without Reliability & 50 & 40 & 40 & 60 \\
 & GrASPE(ours) & \textbf{70} & \textbf{70} & \textbf{80} & \textbf{80} \\
\hline

\multirow{7}{*}{\rotatebox[origin=c]{0}{\makecell{\textbf{Norm.}\\\textbf{Traj.}\\\textbf{Length}}}} 
 & DWA \cite{fox1997dwa}  & 0.38 & 0.56 & 0.33 & 1.32   \\
  & GA-Nav \cite{ganav} & 0.35 & 0.59 & 0.52 & 0.49 \\
  & NMFNet \cite{nguyen2020nmfnet}  & 0.69 & 0.58 & 0.53 & 0.72 \\
 & PAAD \cite{ji2022proactive_ad}  & 0.81 & 0.95 & 0.76 & 0.93 \\
 & GrASPE without A-GNN & 0.42 & 0.63 & 0.38 & 0.43 \\
 & GrASPE without Reliability & 0.89 & 1.25 & 0.96 & 0.77 \\
 & GrASPE(ours) & 1.11 & 1.09 & 1.15 & 1.23 \\
\hline

\multirow{7}{*}{\rotatebox[origin=c]{0}{\makecell{\textbf{ False}\\\textbf{Positive }\\\textbf{Rate}}}} 
& DWA \cite{fox1997dwa} & - & - & - & -   \\
 & GA-Nav \cite{ganav} & 0.33 & 0.39 & 0.30 & 0.61 \\
  & NMFNet \cite{nguyen2020nmfnet}  & 0.38 & 0.41 & 0.56 & 0.59 \\
 & PAAD \cite{ji2022proactive_ad}  & 0.21 & 0.29 & 0.32 & 0.42 \\
  & GrASPE without A-GNN & 0.42 & 0.58 & 0.53 & 0.55\\
 & GrASPE without Reliability & 0.28 & 0.31 & 0.25 & 0.36 \\
 & GrASPE(ours) &  \textbf{0.15} &  \textbf{0.18} &  \textbf{0.12} &  \textbf{0.21} \\
\hline

\multirow{7}{*}{\rotatebox[origin=c]{0}{\makecell{\textbf{ False}\\\textbf{Negative }\\\textbf{Rate}}}} 
& DWA \cite{fox1997dwa} & - & - & - & -   \\
 & GA-Nav \cite{ganav} & 0.41 & 0.34 & 0.46 & 0.58 \\
 & NMFNet \cite{nguyen2020nmfnet}  & 0.46 & 0.38 & 0.49 & 0.57 \\
 & PAAD \cite{ji2022proactive_ad}  & 0.34 & 0.23 & 0.31 & 0.39 \\
  & GrASPE without A-GNN & 0.48 & 0.55 & 0.59 & 0.61 \\
 & GrASPE without Reliability & 0.22 & 0.25 & 0.29 & 0.19 \\
 & GrASPE(ours) &  \textbf{0.09} &  \textbf{0.16} &  \textbf{0.17} &  \textbf{0.19} \\
\hline

\end{tabular}
}
\caption{\small{Navigation performance of our method compared to other methods on various metrics. GrASPE outperforms other methods consistently in terms of success rate, false positive rate, and false negative rate in different unstructured outdoor scenarios including low-light conditions.  }
}
\label{tab:comparison_table}
\vspace{-17pt}
\end{table}

\begin{table}
\begin{center}
\small
\begin{tabular}{ |c|c|c|c| } 
\hline
Method & Inference Rate (Hz)  \\
\hline
\multirow{1}{12em}{GA-Nav \cite{ganav}} & 14.832 \\ 
\multirow{1}{12em}{NMFNet \cite{nguyen2020nmfnet}} & 10.725  \\ 
\multirow{1}{12em}{PAAD \cite{ji2022proactive_ad}} & 13.578  \\ 
\multirow{1}{12em}{GrASPE without A-GNN} & 19.415  \\ 
\multirow{1}{12em}{GrASPE without Reliability} & 9.167  \\ 
\multirow{1}{12em}{GrASPE (ours)} & 8.873  \\ 

\hline
\end{tabular}
\end{center}
\caption{\small Inference rates (higher is better) of the methods in comparisons. GrASPE has a low inference rate due to the computation-heavy nature of GNNs. However, we note that to the best of our knowledge, GrASPE is the first GNN-based formulation that executes in real-time which is a prerequisite for robot navigation.}
\vspace{-17pt}
\label{tab:inference_table}
\end{table}

\section{Acknowledgement}
This work was supported in part by ARO Grants W911NF2110026,  and Army Cooperative Agreement W911NF2120076. We acknowledge the support of the Maryland Robotics Center.

\section{Conclusions, Limitations and Future Work}

We present a novel multi-modal fusion algorithm to navigate a legged robot in unstructured outdoor environments where the sensors experience distortions. We utilize a graph attention-based prediction model along with a sensor reliability estimator to obtain a given trajectory's navigation success probabilities. This prediction model is combined with a local planner to generate navigation actions while avoiding actions corresponding to unsuccessful trajectories from the prediction model. We validate our method's performance in different unstructured environments and compare it with the other methods qualitatively and quantitatively.

Our algorithm has a few limitations. The method assumes non-holonomic robot dynamics to ensure that the trajectories can be projected into the RGB image. However, the legged robot dynamics are holonomic and further investigation is required to extend our fusion strategy to relax the action constraints. Our method could cause collisions in extreme cases where all the perception sensors become unreliable. Adding a haptic sensor modality to navigate more cautiously could help reduce the impact of such failures. A new sensor modality can be easily added to our pipeline to enhance GrASPE's capabilities. However, its effect on the real-time execution needs to be analyzed.

\bibliographystyle{IEEEtran}
\bibliography{References}

\begin{thebibliography}{10}
\providecommand{\url}[1]{#1}
\csname url@samestyle\endcsname
\providecommand{\newblock}{\relax}
\providecommand{\bibinfo}[2]{#2}
\providecommand{\BIBentrySTDinterwordspacing}{\spaceskip=0pt\relax}
\providecommand{\BIBentryALTinterwordstretchfactor}{4}
\providecommand{\BIBentryALTinterwordspacing}{\spaceskip=\fontdimen2\font plus
\BIBentryALTinterwordstretchfactor\fontdimen3\font minus
  \fontdimen4\font\relax}
\providecommand{\BIBforeignlanguage}[2]{{%
\expandafter\ifx\csname l@#1\endcsname\relax
\typeout{** WARNING: IEEEtran.bst: No hyphenation pattern has been}%
\typeout{** loaded for the language `#1'. Using the pattern for}%
\typeout{** the default language instead.}%
\else
\language=\csname l@#1\endcsname
\fi
#2}}
\providecommand{\BIBdecl}{\relax}
\BIBdecl

\bibitem{limosani2018delivery}
R.~Limosani, R.~Esposito, A.~Manzi, G.~Teti, F.~Cavallo, and P.~Dario,
  ``Robotic delivery service in combined outdoor--indoor environments:
  technical analysis and user evaluation,'' \emph{Robotics and autonomous
  systems}, vol. 103, pp. 56--67, 2018.

\bibitem{roldan2018agriculture}
J.~J. Rold{\'a}n, J.~del Cerro, D.~Garz{\'o}n-Ramos, P.~Garcia-Aunon,
  M.~Garz{\'o}n, J.~De~Le{\'o}n, and A.~Barrientos, ``Robots in agriculture:
  State of art and practical experiences,'' \emph{Service robots}, pp. 67--90,
  2018.

\bibitem{zaheer2021surveillance}
M.~Z. Zaheer, A.~Mahmood, M.~H. Khan, M.~Astrid, and S.-I. Lee, ``An anomaly
  detection system via moving surveillance robots with human collaboration,''
  in \emph{Proceedings of the IEEE/CVF International Conference on Computer
  Vision}, 2021, pp. 2595--2601.

\bibitem{gu2018exploration}
Y.~Gu, J.~Strader, N.~Ohi, S.~Harper, K.~Lassak, C.~Yang, L.~Kogan, B.~Hu,
  M.~Gramlich, R.~Kavi \emph{et~al.}, ``Robot foraging: Autonomous sample
  return in a large outdoor environment,'' \emph{IEEE Robotics \& Automation
  Magazine}, vol.~25, no.~3, pp. 93--101, 2018.

\bibitem{choi2019rescue}
B.~Choi, W.~Lee, G.~Park, Y.~Lee, J.~Min, and S.~Hong, ``Development and
  control of a military rescue robot for casualty extraction task,''
  \emph{Journal of Field Robotics}, vol.~36, no.~4, pp. 656--676, 2019.

\bibitem{aladem2019low_light_nav}
M.~Aladem, S.~Baek, and S.~A. Rawashdeh, ``Evaluation of image enhancement
  techniques for vision-based navigation under low illumination,''
  \emph{Journal of Robotics}, vol. 2019, 2019.

\bibitem{ji2022proactive_ad}
T.~Ji, A.~N. Sivakumar, G.~Chowdhary, and K.~Driggs-Campbell, ``Proactive
  anomaly detection for robot navigation with multi-sensor fusion,'' \emph{IEEE
  Robotics and Automation Letters}, vol.~7, no.~2, pp. 4975--4982, 2022.

\bibitem{seeing-through-fog}
M.~Bijelic, T.~Gruber, F.~Mannan, F.~Kraus, W.~Ritter, K.~Dietmayer, and
  F.~Heide, ``Seeing through fog without seeing fog: Deep multimodal sensor
  fusion in unseen adverse weather,'' in \emph{The IEEE Conference on Computer
  Vision and Pattern Recognition (CVPR)}, June 2020.

\bibitem{nguyen2020nmfnet}
A.~Nguyen, N.~Nguyen, K.~Tran, E.~Tjiputra, and Q.~D. Tran, ``Autonomous
  navigation in complex environments with deep multimodal fusion network,'' in
  \emph{2020 IEEE/RSJ International Conference on Intelligent Robots and
  Systems (IROS)}.\hskip 1em plus 0.5em minus 0.4em\relax IEEE, 2020, pp.
  5824--5830.

\bibitem{fox1997dwa}
D.~Fox, W.~Burgard, and S.~Thrun, ``The dynamic window approach to collision
  avoidance,'' \emph{IEEE Robotics \& Automation Magazine}, vol.~4, no.~1, pp.
  23--33, 1997.

\bibitem{sathyamoorthy2022terrapn}
A.~J. Sathyamoorthy, K.~Weerakoon, T.~Guan, J.~Liang, and D.~Manocha,
  ``Terrapn: Unstructured terrain navigation through online self-supervised
  learning,'' \emph{arXiv preprint arXiv:2202.12873}, 2022.

\bibitem{mansouri2020_lidar_camera}
S.~S. Mansouri, C.~Kanellakis, D.~Kominiak, and G.~Nikolakopoulos, ``Deploying
  mavs for autonomous navigation in dark underground mine environments,''
  \emph{Robotics and Autonomous Systems}, vol. 126, p. 103472, 2020.

\bibitem{terp}
K.~Weerakoon, A.~J. Sathyamoorthy, U.~Patel, and D.~Manocha, ``Terp: Reliable
  planning in uneven outdoor environments using deep reinforcement learning,''
  in \emph{2022 International Conference on Robotics and Automation (ICRA)},
  2022, pp. 9447--9453.

\bibitem{weerakoon2022htron}
K.~Weerakoon, S.~Chakraborty, N.~Karapetyan, A.~J. Sathyamoorthy, A.~S. Bedi,
  and D.~Manocha, ``Htron: Efficient outdoor navigation with sparse rewards via
  heavy tailed adaptive reinforce algorithm,'' \emph{arXiv preprint
  arXiv:2207.03694}, 2022.

\bibitem{cai2020probabilistic_camera_lidar}
P.~Cai, S.~Wang, Y.~Sun, and M.~Liu, ``Probabilistic end-to-end vehicle
  navigation in complex dynamic environments with multimodal sensor fusion,''
  \emph{IEEE Robotics and Automation Letters}, vol.~5, no.~3, pp. 4218--4224,
  2020.

\bibitem{debeunne2020review_camra_lidar}
C.~Debeunne and D.~Vivet, ``A review of visual-lidar fusion based simultaneous
  localization and mapping,'' \emph{Sensors}, vol.~20, no.~7, p. 2068, 2020.

\bibitem{zhang2019gnn_vs_cnn}
S.~Zhang, H.~Tong, J.~Xu, and R.~Maciejewski, ``Graph convolutional networks: a
  comprehensive review,'' \emph{Computational Social Networks}, vol.~6, no.~1,
  pp. 1--23, 2019.

\bibitem{wu2020gnn_review}
Z.~Wu, S.~Pan, F.~Chen, G.~Long, C.~Zhang, and S.~Y. Philip, ``A comprehensive
  survey on graph neural networks,'' \emph{IEEE transactions on neural networks
  and learning systems}, vol.~32, no.~1, pp. 4--24, 2020.

\bibitem{gnn_fusion}
S.~Casas, C.~Gulino, R.~Liao, and R.~Urtasun, ``Spagnn: Spatially-aware graph
  neural networks for relational behavior forecasting from sensor data,'' in
  \emph{2020 IEEE International Conference on Robotics and Automation (ICRA)},
  2020, pp. 9491--9497.

\bibitem{ravichandran2022gnn_indoor_nav}
Z.~Ravichandran, L.~Peng, N.~Hughes, J.~D. Griffith, and L.~Carlone,
  ``Hierarchical representations and explicit memory: Learning effective
  navigation policies on 3d scene graphs using graph neural networks,'' in
  \emph{2022 International Conference on Robotics and Automation (ICRA)}.\hskip
  1em plus 0.5em minus 0.4em\relax IEEE, 2022, pp. 9272--9279.

\bibitem{uncertainity_aware_nav}
K.~Katyal, K.~Popek, C.~Paxton, P.~Burlina, and G.~D. Hager,
  ``Uncertainty-aware occupancy map prediction using generative networks for
  robot navigation,'' in \emph{2019 International Conference on Robotics and
  Automation (ICRA)}, 2019, pp. 5453--5459.

\bibitem{rusli2021vision_nav}
L.~Rusli, B.~Nurhalim, and R.~Rusyadi, ``Vision-based vanishing point detection
  of autonomous navigation of mobile robot for outdoor applications,''
  \emph{Journal of Mechatronics, Electrical Power, and Vehicular Technology},
  vol.~12, no.~2, pp. 117--125, 2021.

\bibitem{slip_aware_fusion}
F.~Liu, X.~Li, S.~Yuan, and W.~Lan, ``Slip-aware motion estimation for off-road
  mobile robots via multi-innovation unscented kalman filter,'' \emph{IEEE
  Access}, vol.~8, pp. 43\,482--43\,496, 2020.

\bibitem{yan2022multi_sensor_ekf}
Y.~Yan, B.~Zhang, J.~Zhou, Y.~Zhang, X.~Liu \emph{et~al.}, ``Real-time
  localization and mapping utilizing multi-sensor fusion and visual--imu--wheel
  odometry for agricultural robots in unstructured, dynamic and gps-denied
  greenhouse environments,'' \emph{Agronomy}, vol.~12, no.~8, p. 1740, 2022.

\bibitem{qu2021multi_sensor_indoor}
Y.~Qu, M.~Yang, J.~Zhang, W.~Xie, B.~Qiang, and J.~Chen, ``An outline of
  multi-sensor fusion methods for mobile agents indoor navigation,''
  \emph{Sensors}, vol.~21, no.~5, p. 1605, 2021.

\bibitem{bishop2001kalman}
G.~Bishop, G.~Welch \emph{et~al.}, ``An introduction to the kalman filter,''
  \emph{Proc of SIGGRAPH, Course}, vol.~8, no. 27599-23175, p.~41, 2001.

\bibitem{ekf_1}
A.~E.~B. Velasquez, V.~A.~H. Higuti, M.~V. Gasparino, A.~N. Sivakumar,
  M.~Becker, and G.~Chowdhary, ``Multi-sensor fusion based robust row following
  for compact agricultural robots,'' \emph{arXiv preprint arXiv:2106.15029},
  2021.

\bibitem{su2021_lidar_imu}
Y.~Su, T.~Wang, S.~Shao, C.~Yao, and Z.~Wang, ``Gr-loam: Lidar-based sensor
  fusion slam for ground robots on complex terrain,'' \emph{Robotics and
  Autonomous Systems}, vol. 140, p. 103759, 2021.

\bibitem{liang2022adaptiveon}
J.~Liang, K.~Weerakoon, T.~Guan, N.~Karapetyan, and D.~Manocha, ``Adaptiveon:
  Adaptive outdoor navigation method for stable and reliable motions,''
  \emph{arXiv preprint arXiv:2205.03517}, 2022.

\bibitem{shin2018slam}
Y.-S. Shin, Y.~S. Park, and A.~Kim, ``Direct visual slam using sparse depth for
  camera-lidar system,'' in \emph{2018 IEEE International Conference on
  Robotics and Automation (ICRA)}.\hskip 1em plus 0.5em minus 0.4em\relax IEEE,
  2018, pp. 5144--5151.

\bibitem{ad_reactive1_ji2021}
T.~Ji, S.~T. Vuppala, G.~Chowdhary, and K.~Driggs-Campbell, ``Multi-modal
  anomaly detection for unstructured and uncertain environments,'' in
  \emph{Conference on Robot Learning}.\hskip 1em plus 0.5em minus 0.4em\relax
  PMLR, 2021, pp. 1443--1455.

\bibitem{ad_reactive_2_park2019}
D.~Park, H.~Kim, and C.~C. Kemp, ``Multimodal anomaly detection for assistive
  robots,'' \emph{Autonomous Robots}, vol.~43, no.~3, pp. 611--629, 2019.

\bibitem{zhou2012self_multi_sensor}
S.~Zhou, J.~Xi, M.~W. McDaniel, T.~Nishihata, P.~Salesses, and K.~Iagnemma,
  ``Self-supervised learning to visually detect terrain surfaces for autonomous
  robots operating in forested terrain,'' \emph{Journal of Field Robotics},
  vol.~29, no.~2, pp. 277--297, 2012.

\bibitem{multimodal_HMM_anomaly}
D.~Park, H.~Kim, and C.~C. Kemp, ``Multimodal anomaly detection for assistive
  robots,'' \emph{Autonomous Robots}, vol.~43, no.~3, pp. 611--629, 2019.

\bibitem{ad_multimodal_wellhausen2020}
L.~Wellhausen, R.~Ranftl, and M.~Hutter, ``Safe robot navigation via
  multi-modal anomaly detection,'' \emph{IEEE Robotics and Automation Letters},
  vol.~5, no.~2, pp. 1326--1333, 2020.

\bibitem{kahn2021badgr}
G.~Kahn, P.~Abbeel, and S.~Levine, ``Badgr: An autonomous self-supervised
  learning-based navigation system,'' \emph{IEEE Robotics and Automation
  Letters}, vol.~6, no.~2, pp. 1312--1319, 2021.

\bibitem{jin2022gnn_anomaly}
K.~Jin, H.~Wang, C.~Liu, Y.~Zhai, and L.~Tang, ``Graph neural network based
  relation learning for abnormal perception information detection in
  self-driving scenarios,'' in \emph{2022 International Conference on Robotics
  and Automation (ICRA)}.\hskip 1em plus 0.5em minus 0.4em\relax IEEE, 2022,
  pp. 8943--8949.

\bibitem{sorokin2022learning}
M.~Sorokin, J.~Tan, C.~K. Liu, and S.~Ha, ``Learning to navigate sidewalks in
  outdoor environments,'' \emph{IEEE Robotics and Automation Letters}, vol.~7,
  no.~2, pp. 3906--3913, 2022.

\bibitem{ji2021multi_anomaly}
T.~Ji, S.~T. Vuppala, G.~Chowdhary, and K.~Driggs-Campbell, ``Multi-modal
  anomaly detection for unstructured and uncertain environments,'' in
  \emph{Conference on Robot Learning}.\hskip 1em plus 0.5em minus 0.4em\relax
  PMLR, 2021, pp. 1443--1455.

\bibitem{kahn2021land}
G.~Kahn, P.~Abbeel, and S.~Levine, ``Land: Learning to navigate from
  disengagements,'' \emph{IEEE Robotics and Automation Letters}, vol.~6, no.~2,
  pp. 1872--1879, 2021.

\bibitem{gcn}
M.~Welling and T.~N. Kipf, ``Semi-supervised classification with graph
  convolutional networks,'' in \emph{J. International Conference on Learning
  Representations (ICLR 2017)}, 2016.

\bibitem{agcn}
K.~K. Thekumparampil, C.~Wang, S.~Oh, and L.-J. Li, ``Attention-based graph
  neural network for semi-supervised learning,'' \emph{arXiv preprint
  arXiv:1803.03735}, 2018.

\bibitem{rgat}
D.~Busbridge, D.~Sherburn, P.~Cavallo, and N.~Y. Hammerla, ``Relational graph
  attention networks,'' \emph{arXiv preprint arXiv:1904.05811}, 2019.

\bibitem{resnet}
K.~He, X.~Zhang, S.~Ren, and J.~Sun, ``Deep residual learning for image
  recognition,'' in \emph{Proceedings of the IEEE conference on computer vision
  and pattern recognition}, 2016, pp. 770--778.

\bibitem{woo2018cbam}
S.~Woo, J.~Park, J.-Y. Lee, and I.~S. Kweon, ``Cbam: Convolutional block
  attention module,'' in \emph{Proceedings of the European conference on
  computer vision (ECCV)}, 2018, pp. 3--19.

\bibitem{qi2017pointnet}
C.~R. Qi, H.~Su, K.~Mo, and L.~J. Guibas, ``Pointnet: Deep learning on point
  sets for 3d classification and segmentation,'' in \emph{Proceedings of the
  IEEE conference on computer vision and pattern recognition}, 2017, pp.
  652--660.

\bibitem{fast2}
E.~Rosten, R.~Porter, and T.~Drummond, ``Faster and better: A machine learning
  approach to corner detection,'' \emph{IEEE transactions on pattern analysis
  and machine intelligence}, vol.~32, no.~1, pp. 105--119, 2008.

\bibitem{zhang2014loam}
J.~Zhang and S.~Singh, ``Loam: Lidar odometry and mapping in real-time.'' in
  \emph{Robotics: Science and Systems}, vol.~2, no.~9.\hskip 1em plus 0.5em
  minus 0.4em\relax Berkeley, CA, 2014, pp. 1--9.

\bibitem{mojsilovic2002isee}
A.~Mojsilovic, J.~Gomes, and B.~E. Rogowitz, ``Isee: Perceptual features for
  image library navigation,'' in \emph{Human Vision and Electronic Imaging
  VII}, vol. 4662.\hskip 1em plus 0.5em minus 0.4em\relax SPIE, 2002, pp.
  266--277.

\bibitem{fast}
E.~Rosten and T.~Drummond, ``Machine learning for high-speed corner
  detection,'' in \emph{European conference on computer vision}.\hskip 1em plus
  0.5em minus 0.4em\relax Springer, 2006, pp. 430--443.

\bibitem{spectral_theorem_Hall}
B.~C. Hall, \emph{Quantum Theory for Mathematicians}, ser. Graduate Texts in
  Mathematics.\hskip 1em plus 0.5em minus 0.4em\relax Springer New York, no.
  267.

\bibitem{ganav}
T.~Guan, D.~Kothandaraman, R.~Chandra, A.~J. Sathyamoorthy, K.~Weerakoon, and
  D.~Manocha, ``Ga-nav: Efficient terrain segmentation for robot navigation in
  unstructured outdoor environments,'' \emph{IEEE Robotics and Automation
  Letters}, vol.~7, no.~3, pp. 8138--8145, 2022.

\bibitem{pspnet}
H.~Zhao, J.~Shi, X.~Qi, X.~Wang, and J.~Jia, ``Pyramid scene parsing network,''
  in \emph{2017 IEEE Conference on Computer Vision and Pattern Recognition
  (CVPR)}, 2017, pp. 6230--6239.

\bibitem{RUGD2019IROS}
M.~Wigness, S.~Eum, J.~G. Rogers, D.~Han, and H.~Kwon, ``A rugd dataset for
  autonomous navigation and visual perception in unstructured outdoor
  environments,'' in \emph{International Conference on Intelligent Robots and
  Systems (IROS)}, 2019.

\bibitem{jiang2020rellis3d}
P.~Jiang, P.~Osteen, M.~Wigness, and S.~Saripalli, ``Rellis-3d dataset: Data,
  benchmarks and analysis,'' 2020.

\end{thebibliography}

\clearpage
\onecolumn
\section{Appendix} \label{sec:appendix}

\subsection{Proof of Lemma \ref{lem:laplacian}:}  

From Graph theory, we consider the graph Laplacian $\mathcal{L} = \mathcal{D} -\mathcal{W}$, where $\mathcal{D}$ is a diagonal matrix with $\mathcal{D}_{i,i} = deg(v_i)$. Here, $deg(v_i)$ is the degree of a vertex which is a measure of the number of edges terminating at that vertex. In this context, we consider $\mathcal{D}_{i,i} = \sum_j w_{i,j}$. By construction and from Lemma \ref{lem:undirected_graph}, we can observe that $\mathcal{D}$ is real and $\mathcal{W}$ is hermitian (i.e., $w_{i,j} = w_{j,i}^*$).

Lemma \ref{lem:laplacian}: Laplacian matrix $\mathcal{L}$ of the graph $\mathcal{G}_t$ is spectrally decomposable.

\begin{proof}
From Lemma \ref{lem:undirected_graph}, $ Re(w_{i,j}) \geq 0 \, \forall \, i,j$. Consider a real valued $x$,\\
\begin{equation}
 \begin{aligned} 
 x^T\mathcal{L}x 
 &= \sum_{i,j} w_{i,j} (x_i-x_j)^2 \quad \because \mathcal{L} =\mathcal{D} -\mathcal{W}\\
 &= \sum_{i < j} w_{i,j} (x_i-x_j)^2 +\sum_{i>j} w_{i,j} (x_i-x_j)^2\\
 &= \sum_{i < j} w_{i,j} (x_i-x_j)^2 +\sum_{i>j} w_{j,i}^* (x_i-x_j)^2\\
&= \sum_{i < j} w_{i,j} (x_i-x_j)^2 +\sum_{i<j} w_{i,j}^* (x_j-x_i)^2\\
&= \sum_{i < j} w_{i,j} (x_i-x_j)^2 +\sum_{i<j} w_{i,j}^* (x_i-x_j)^2\\
&= \sum_{i < j} (w_{i,j} + w_{i,j}^*)(x_i-x_j)^2 \\
&= 2\sum_{i < j}Re (w_{i,j})(x_i-x_j)^2  \geq 0 \\
&\implies \mathcal{L} \quad \textrm{is Positive semi-definite.}
\end{aligned}
\vspace{-2pt}
\end{equation}
$\therefore$  Symmetric matrix $\mathcal{L}$ has all real eigenvalues. Further, the corresponding eigenvectors $u_1,..,u_N$ can be taken to be orthonormal by,
\vspace{-5pt}
\begin{equation}
u_i^T u_j = \begin{cases}
    1,& \text{if } i = j\\
    0,    & \text{if } i\neq j,
\end{cases}
\vspace{-2pt}
\end{equation}

\no Therefore, from the Spectral theorem \cite{spectral_theorem_Hall}, $\mathcal{L}$ can be spectrally decomposed as, $\mathcal{L} = U \Lambda U^T$, where 
$U$ is the eigenvector matrix, and $\Lambda$ denotes the diagonal matrix of sorted eigenvalues.

\end{proof}

\subsection{Data Collection Process (extending Section V.B)} 
The multi-modal dataset used in this work is collected by operating the Spot robot under different lighting conditions in an outdoor field that includes bushes, small trees, hanging leaves, and grass regions of different heights and densities. We chose 8 different locations in the outdoor field and collected data at least for 6 runs including high, medium, and low lighting conditions. The data streams were collected at around 10–15Hz rate since sensors such as LiDAR can provide data only up to 20Hz even though camera images can be collected at 30Hz. The total duration of data collection was ~90 minutes. 

\subsection{Training Setup} 

GrASPE is implemented using PyTorch and PyTorch Geometric (PyG). The prediction model is trained in a workstation with an Intel Xeon 3.6 GHz processor and an Nvidia Titan GPU using real-world data collected from a Boston Dynamics Spot robot.

\begin{table}[h]
\centering
\normalsize
\begin{tabular}{|p{5cm}|p{5cm}|} 
\hline
\textbf{Optimizer} & Adam  \\ 
\hline
\textbf{Learning rate}  & 0.002  \\
\hline
\textbf{Weight decay}  & 0.00025  \\
\hline
\textbf{Epochs}  & 1000  \\
\hline
\textbf{Loss Function}  & Mean squared error (MSE) loss  \\
\hline
\end{tabular}

\caption{\label{Tab:Results2}}
\end{table}

\subsection{Parameters and default values} 

\begin{table}[h]
\centering
\normalsize
\begin{tabular}{|p{2cm}|p{3cm}|p{2cm}|p{2cm}|p{6cm}|} 
\hline
\textbf{Parameter} & \textbf{Description} & \textbf{Range} & \textbf{Default Value} & \textbf{Notes}   \\ [0.5ex] 
\hline
$w$  & Image width & $\mathbb{Z}^+$ & 320 & High-resolution images lead to challenges in real-time performance. \\
\hline
$h$  & Image height & $\mathbb{Z}^+$ & 240 &  \\
\hline
$N_p$  & Point cloud length & $\mathbb{Z}^+$ & 10000 &  Larger point cloud sizes are computationally expensive to process using PointNet. \\
\hline
$N$  & Feature vector $f_{vec}$ length & $\mathbb{Z}^+$ & 120 &  Empirically chosen to reduce the complexity of the graph while maintaining reasonable prediction accuracy. Individual feature vector lengths for $f_{img},f_{point},f_{vel},f_{traj}$ are chosen accordingly to fit the size $N$. \\
\hline
$T$  & Time horizon/ success probability vector length & $\mathbb{Z}^+$ & 10 &  When the success probability vector length is higher, the labeling requires more effort, and higher dimensional predictions lead to lower prediction accuracy. \\
\hline
$d_{th}$  & Goal reaching threshold & $\mathbb{R}^+$ & 0.5 &  Higher threshold will indicate goal reaching before the robot actually reaches the goal. Lower values will lead to circular trajectories near the goal position due to slight errors in the localization. \\
\hline
$\alpha_b$  & Image brightness weighting parameter & $[0,1]$ & 0.67 &  Higher values impose a bias towards image brightness in the overall image reliability estimation $r_{img}$. \\
\hline
$\alpha_c$  & Image corner weighting parameter & $[0,1]$ & 0.33 &  \\
\hline
$C_{max}$  & Maximum smoothness threshold & $[0,1]$ & 0.65 &  Higher values can disregard some edge features in the point cloud. \\
\hline
$C_{min}$  & Minimum smoothness threshold & $[0,1]$ & 0.35 &  Higher values could consider some edge features also as planar features in the point cloud. \\
\hline
$r_{th}$  & Pointcloud reliability threshold & $[0,1]$ & 0.4 &  Lower values could consider unreliable point cloud observations to calculate the obstacle space. This could lead to robot freezing and goal-reaching failures.  \\
\hline
$\beta_e$  & Pointcloud edges weighting parameter & $[0,1]$  & 0.55 &  Higher values impose a bias towards edge feature richness in the overall point cloud reliability estimation $r_{point}$. \\
\hline
$\beta_p$  & Pointcloud planar feature weighting parameter & $[0,1]$  & 0.45 &   \\
\hline
$\lambda$  & Graph edge weight scalar & $\mathbb{R}^+$  & 15 &  Higher values lead to steep weight decays. The edge weight difference can be unnoticeable for slight changes in node value difference and reliability. \\
\hline
$\gamma_1$  & Goal heading cost weight  & $\mathbb{R}^+$  & 2.4 &  Higher values lead to straight trajectories s.t. The robot attempts to move towards the goal in a straight path. \\
\hline
$\gamma_2$  & Robot velocity cost weight   & $\mathbb{R}^+$  & 0.5 &  Higher values will choose high linear and angular velocities for navigation. Sometimes faster robot motion could lead to failures due to insufficient reaction time. \\
\hline
$\gamma_3$  & Obstacle cost weight   & $\mathbb{R}^+$  & 3.2 &  Higher values will plan actions closer to the obstacles. Can cause collisions in situations where the free spaces are narrow. \\
\hline
$e_{th}$  & Success probability threshold   & $[0,1]$  & 0.56 & Empirically chosen after testing in real-world environments.   \\
\hline
\end{tabular}

\caption{\label{Tab:Results2}}
\end{table}

\end{document}